%% file: main.tex
\documentclass[runningheads]{tpls/llncs}
\usepackage{graphicx}

\usepackage{hyperref}
\usepackage[table]{xcolor}
\usepackage{tikz}
\usepackage{comment}
\usepackage{amsmath,amssymb}
\usepackage{color}
\usepackage{orcidlink}

\usepackage[accsupp]{axessibility}  

\usepackage{style}
\usepackage{style_suppmat}
\usepackage{glossaries}
\usepackage{adjustbox}
\usepackage{tikz-3dplot}
\usepackage{pgfplots}
\usepackage{xcolor,pgf}
\usepackage[most]{tcolorbox}
\usepackage{caption}
\usepackage{subcaption}
\usepackage{graphicx}
\usepackage{amsmath}
\usepackage{amssymb}
\usepackage{booktabs}
\usepackage{balance}
\usepackage{array}
\usepackage{multirow}
\usepackage[]{algorithm2e}
\usepackage{etoolbox}
\usepackage{xfrac}
\usepackage{enumitem}
\usepackage[normalem]{ulem}
\usepackage{placeins}
\usepackage{pifont}
\usepackage{diagbox}
\usepackage{tabularx}
\usepackage{microtype}
\usepackage{collcell}
\usepackage{hhline}

\usetikzlibrary{shapes.geometric, shadings}
\usepgfplotslibrary{colorbrewer}

\pgfplotsset{compat=1.3}

\begin{document}
\pagestyle{headings}
\mainmatter

\title{Online Segmentation of LiDAR Sequences: Dataset and Algorithm} 

\titlerunning{Online Segmentation of LiDAR Sequences: Dataset and Algorithm}

\author{Romain Loiseau\inst{1,2}\orcidlink{0000-0002-1804-0735}
\and
Mathieu Aubry\inst{1}\orcidlink{0000-0002-3804-0193}
\and
Loïc Landrieu\inst{2}\orcidlink{0000-0002-7738-8141}
}
\authorrunning{R. Loiseau \etal}

\institute{LIGM, Ecole des Ponts, Univ Gustave Eiffel, CNRS, France
\and LASTIG, Univ Gustave Eiffel, IGN, ENSG}

\maketitle

\begin{abstract}
\input{text/FINAL/00_abstract}
\end{abstract}

\captionsetup[table]{font=small}
\captionsetup[figure]{font=small}
\captionsetup[subfigure]{font=small}

\input{text/FINAL/10_introduction}
\input{text/FINAL/20_related_work}
\input{text/FINAL/30_helixnet}
\input{text/FINAL/40_method}
\input{text/FINAL/50_applications}
\input{text/FINAL/90_conclusion}

\input{text/91_acknowledgements}

\bibliographystyle{tpls/splncs04}
\bibliography{egbib}

\clearpage

\clearpage

\section{Supplementary Material}
\input{text/FINAL/99_suppmat}

\end{document}

%% file: text/FINAL/00_abstract.tex
Roof-mounted spinning LiDAR sensors are widely used by autonomous vehicles. However, most semantic datasets and algorithms used for LiDAR sequence segmentation operate on $360^\circ$ frames, causing an acquisition latency incompatible with real-time applications. To address this issue, we first introduce HelixNet, a $10$ billion point dataset with fine-grained labels, timestamps, and sensor rotation information necessary to accurately assess the real-time readiness of segmentation algorithms. Second, we propose Helix4D, a compact and efficient spatio-temporal transformer architecture specifically designed for rotating LiDAR sequences. Helix4D operates on acquisition slices corresponding to a fraction of a full sensor rotation, significantly reducing the total latency. Helix4D reaches accuracy on par with the best segmentation algorithms on HelixNet and SemanticKITTI with a reduction of over $5\times$ in terms of latency and $50\times$ in model size. The code and data are available at: {\tt{\url{https://romainloiseau.fr/helixnet}}}

%% file: text/FINAL/10_introduction.tex
\section{Introduction}\label{sec:introduction}

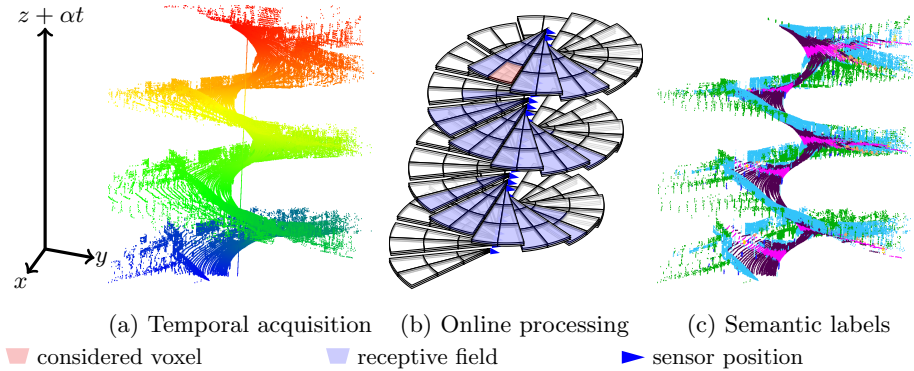
\begin{figure}[t]
    \captionsetup[subfigure]{justification=centering}
    \centering
    \begin{subfigure}[b]{.07\textwidth}  
        \tdplotsetmaincoords{60}{110}
        \begin{tikzpicture}[tdplot_main_coords]
        \draw [very thick, ->] (0, 0, 0) -- (.7, 0, 0);
        \draw [very thick, ->] (0, 0, 0) -- (0, .7, 0); 
        \draw [very thick, ->] (0, 0, 0) -- (0, 0, 3.4);
        \node [draw=none] (nodezt) at (0.0, 0.08, 3.6) {$z + \alpha t$};
        \node [draw=none] (nodex) at (.95, 0, 0) {$x$};
        \node [draw=none] (nodey) at (0, .8, 0){$y$};
        \end{tikzpicture}
    \end{subfigure}
    \hfill
    \begin{subfigure}[t]{.29\textwidth}
        \includegraphics[width=\textwidth, height=.2\textheight]{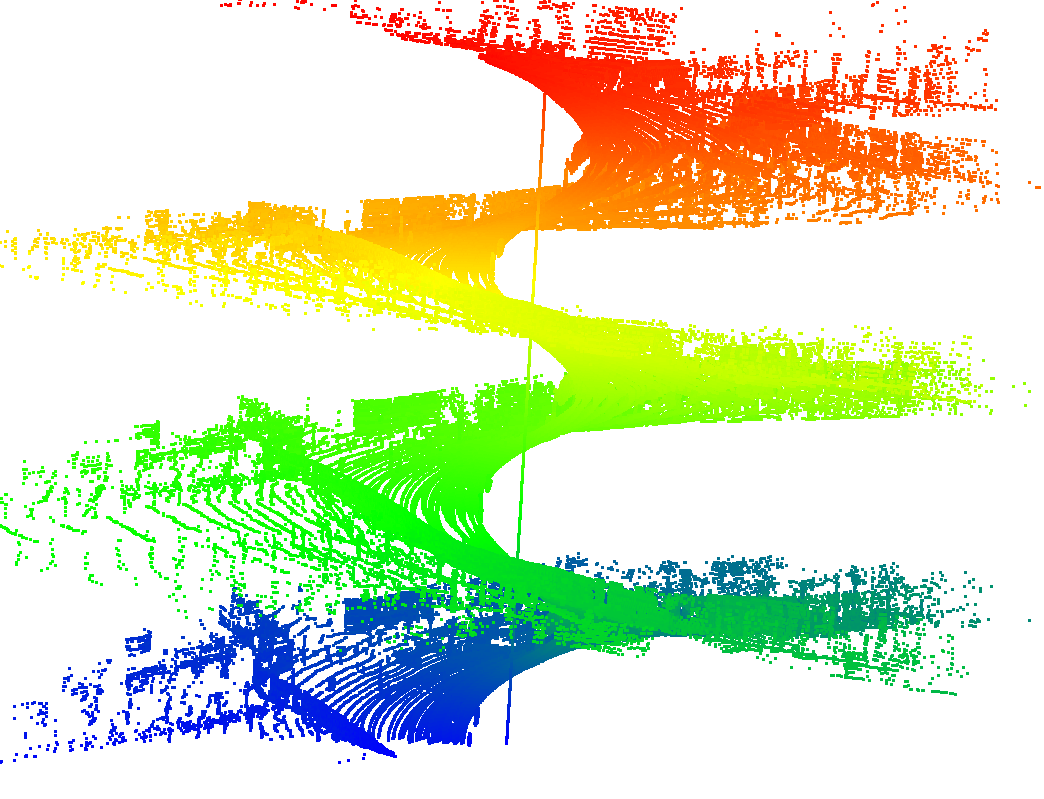}
        \subcaption{\scalebox{1}{Temporal acquisition}}
        \label{fig:helix:time}
    \end{subfigure}
    \begin{subfigure}[t]{.29\textwidth} 
        \includegraphics[width=\textwidth, height=.2\textheight]{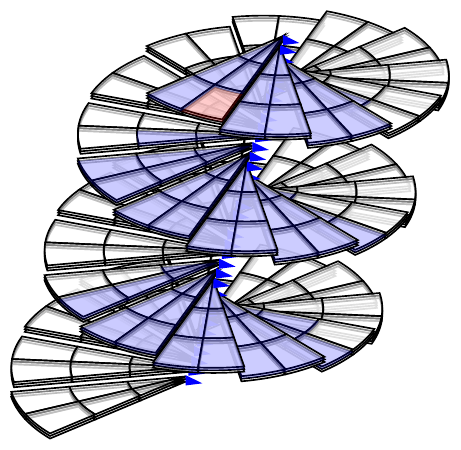}
        \subcaption{\scalebox{1}{Online processing}}
        \label{fig:helix:processing}
    \end{subfigure}
    \begin{subfigure}[t]{.29\textwidth}
        \includegraphics[width=\textwidth, height=.2\textheight]{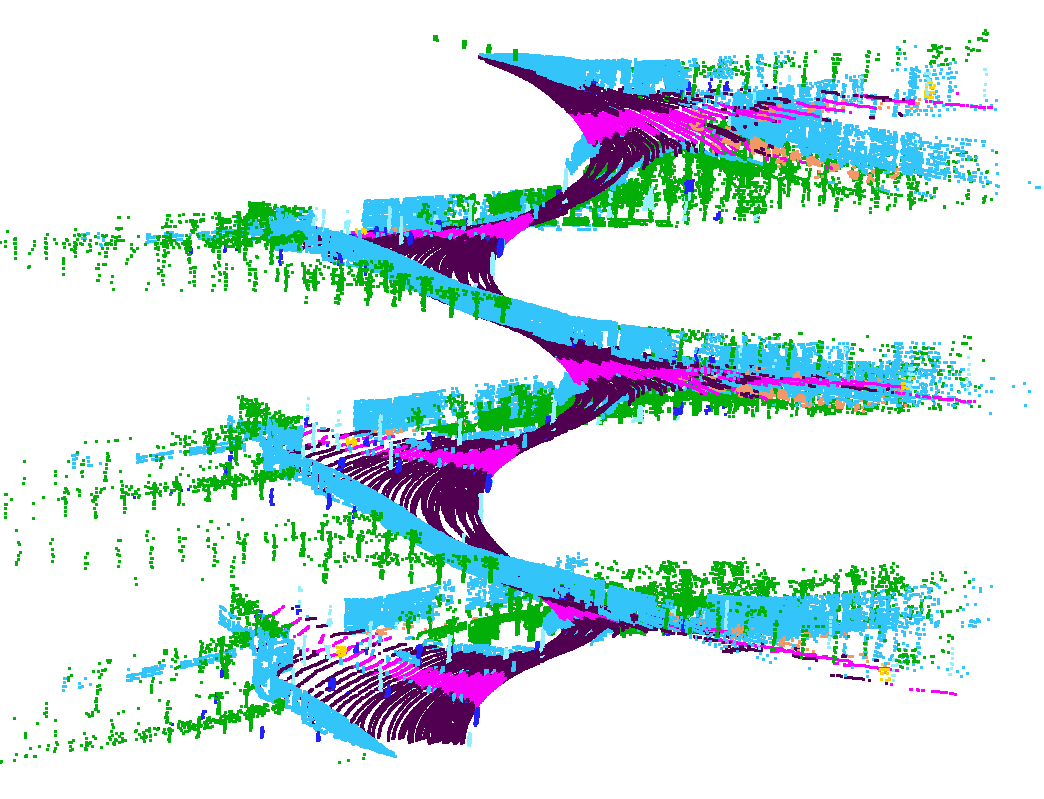}
        \subcaption{\scalebox{1}{{Semantic labels}}}
        \label{fig:helix:sem}
    \end{subfigure}
    
    \begin{tabular*}{\textwidth}{@{}@{\extracolsep{\fill}}*{4}{l}@{}}
    \scalebox{.95}{\intexttrapezium{redvoxel}~considered voxel} &
    \scalebox{.95}{\intexttrapezium{bluevoxel}~receptive field} &
    \scalebox{.95}{\intextsensor~sensor position}
    \end{tabular*}
    \caption{\textbf{Online LiDAR Segmentation}.
    {The 3D point sequences of rotating LiDAR data of our proposed dataset HelixNet follow a complex helix-like structure in space and time, represented in \Subref{fig:helix:time} by using the vertical axis for both time and elevation. We propose an efficient spatio-temporal transformer to process angular slices of data centered on the sensor's position. The slices are partitioned into voxels, each attending other voxels from past slices to build a large spatio-temporal receptive field \Subref{fig:helix:processing}. Our proposed model can segment the LiDAR point stream \Subref{fig:helix:sem} with state-of-the-art accuracy and in real-time.}
    }
    \label{fig:helix}
\end{figure}

Due to their low acquisition latency and high precision, rotating LiDAR sensors are among the most prevalent sensors for autonomous vehicles~\cite{royo2019overview}.
The acquired sequences of 3D points exhibit a complex structure in which the temporal and spatial dimensions are entangled through the rotation of the sensor around a reference point in motion; see \figref{fig:helix}.
However, this structure is often not reflected in the formatting of open-access LiDAR datasets \cite{behley2019iccv,jiang2021rellis,Liao2021ARXIV}, which are discrete sequences of range images, or frames, each corresponding to a 360$^\circ$ degree arc around the sensor.
Consequently, most LiDAR semantic segmentation methods operate on one or several such frames at the same time, in the image \cite{cortinhal2020salsanext} or point cloud \cite{zhu2021cylindrical,Zhang_2020_CVPR,tang2020searching} format.
However, waiting for an entire frame to be acquired introduces an unavoidable latency of more than $100$ms on top of the processing time, excluding applications for high-speed or urban driving. In this paper, we address this issue by introducing (i) HelixNet, the largest available LiDAR dataset, and whose fine-grained point information allows for the realistic real-time evaluation of segmentation methods, and (ii) Helix4D, a spatio-temporal transformer designed for the efficient segmentation of LiDAR sequences.

Our dataset HelixNet, has several key advantages compared to standard datasets such as SemanticKITTI~\cite{behley2019iccv}, see \tabref{tab:helixnetvsothers}. 
By organizing points with respect to sensor rotation and reporting their precise release times, we can accurately benchmark the real-time readiness of leading state-of-the-art LiDAR sequence segmentation algorithms.  
Furthermore, the pointwise sensor orientation allows us to split the data into slices of acquisition corresponding to a fraction of the sensor's rotation. These slices can be processed sequentially by our proposed network Helix4D, resulting in a lower acquisition latency and a more realistic scenario for autonomous driving. 
Based on a spatio-temporal transformer designed explicitly for LiDAR sequences, Helix4D is more than $50$ times smaller than the current best semantic segmentation architectures and reaches state-of-the-art performance with significantly reduced latency.

%% file: text/FINAL/20_related_work.tex
\section{Related work}\label{sec:related_work}

\begin{table}[t]
    \centering
    \caption{{\bf Embarked LiDAR Datasets with Semantic Point Annotations.} 
    With over $8.8$B annotated 3D points, HelixNet is $70$\% larger than SemanticKITTI, and includes more diverse scenes spanning $6$ different French cities. Contrary to other datasets, HelixNet arranges points with respect to the sensor rotation and contains fine-grained information about their release time.}
    \label{tab:helixnetvsothers}
    \begin{tabular*}{\textwidth}{@{}l@{\extracolsep{\fill}}*{5}{l}@{}}\toprule
        \multirow{1}{*}{Dataset} &  
        \multirow{1}{*}{labels} &  \multirow{1}{*}{frames} &  \multirow{1}{*}{classes} &
        \multirow{1}{*}{span} &
        \multirow{1}{*}{format}
        \\
        \midrule
        \bf HelixNet {(Ours)} & {8.85B} & 78k & 9 & {6 cities} & sensor rotation
         \\
        SemanticKITTI~\cite{Geiger2013IJRR,behley2019iccv} & 5.2B & 43k & 19 & 1 city & frame 
        \\
        Rellis3D~\cite{jiang2021rellis} & 1.5B & 13k & 16 & 1 city & frame  
        \\
        KITTI-360~\cite{Liao2021ARXIV}& 1.0B & 81k & 37 & 1 city & frame 
        \\
        A2D2~\cite{geyer2020a2d2} & 387M & 41k & 38 & 3 cities & frames\\
        Paris-Lille-3D \cite{roynard2017parislille3d} & 143M & N/A & 50 & 2 cities & multi-frame
        \\
        Toronto3D \cite{tan2020toronto} & 78M & N/A& 8 & 1 city &  multi-frame 
        \\
        \bottomrule
    \end{tabular*}
\end{table}

We present an overview of the existing LiDAR datasets related to autonomous driving, and a summary of the recent developments in 3D semantic segmentation.

\paragraph{Autonomous Driving 3D Datasets.}
As autonomous driving becomes an increasingly realistic prospect, multiple datasets have been proposed to evaluate the performance of perception algorithms \cite{neuhold2017mapillary,cordts2016cityscapes}. In addition to cameras, rotating LiDARs have become one of the most prevalent sensors mounted on autonomous vehicles due to their high accuracy, low latency, and steadily decreasing prices~\cite{royo2019overview}. ApolloScape \cite{huang2018apolloscape}, DublinCity~\cite{zolanvari2019dublincity}, and TerraMobilita/iQmulus~\cite{vallet2015terramobilita} have been acquired with LiDAR setups that offer scans of urban environments with high precision and density. However, the vertical orientation of the emitters is not compatible with real-time road perception. 
Several prominent datasets such as NuScene~\cite{caesar2020nuscenes} or the very large ONCE dataset~\cite{mao2021one} provide only object-level annotations (\ie boxes). 

This paper focuses on semantic segmentation algorithms for roof-mounted rotating LiDAR sensors. In \tabref{tab:helixnetvsothers}, we report several key characteristics of such datasets~\cite{jiang2021rellis,Geiger2013IJRR,behley2019iccv,Liao2021ARXIV,geyer2020a2d2,roynard2017parislille3d,tan2020toronto}.
Our proposed dataset HelixNet is $70$\% larger than SemanticKITTI \cite{Geiger2013IJRR,behley2019iccv}, and spans $6$ cities and various environments.
In contrast to previously released datasets, the 3D points of HelixNet are given with respect to the sensor rotation and in the order in which they are made available. This last point proves crucial for evaluating the precision and latency of segmentation algorithms in a setting that is compatible with real-time inference.

\paragraph{Deep Semantic Segmentation of 3D Point Clouds.}
The development of specific deep architectures for the semantic segmentation of 3D point clouds has led to a tremendous increase in performance~\cite{guo2020deep}. 
The first set of methods that operate on rotating LiDAR sequences processes data in range image format \cite{Liang_2021_CVPR,Sun_2021_CVPR,cortinhal2020salsanext}.
Taking advantage of advances in the implementation of sparse convolutions~\cite{SubmanifoldSparseConvNet,choy20194d}, a second set of methods uses fine grids in polar~\cite{Zhang_2020_CVPR}, Cartesian \cite{tang2020searching,cheng20212} or cylindrical~\cite{zhu2021cylindrical,Hong_2021_CVPR} coordinates. A third kind of approach proposes exploiting the temporal dimension of LiDAR acquisitions by \emph{stacking} contiguous frames~\cite{choy20194d,aygun20214d}. Observing that cylindrical partitions better capture the geometry of rotating LiDAR acquisition, our proposed Helix4D builds on the idea of Cylinder3D~\cite{zhu2021cylindrical} and adds a temporal component to the architecture.

Due to their remarkable performance and scalability, transformers~\cite{NIPS2017_3f5ee243} have quickly been adapted from text processing to images~\cite{dosovitskiy2020image,Liu_2021_ICCV,Strudel_2021_ICCV,carion2020end}, videos~\cite{Arnab_2021_ICCV}, or meshes~\cite{lin2021end}. Transformers are also well suited to handle unordered sets, such as 3D point clouds~\cite{guo2021pct,zhao2021point}. 
In particular, their scalability can be leveraged to achieve large receptive fields~\cite{Mao_2021_ICCV,pan20213d} and more discriminative features~\cite{bhattacharyya2021self} than purely convolutional approaches. Transformers can also efficiently process complex temporal~\cite{katharopoulos_et_al_2020,vyas_et_al_2020} and spatio-temporal~\cite{fan2021point} sequences. In the wake of hybrid convolution-transformer models~\cite{guo2021cmt,coccomini2021combining,d2021convit}, our proposed Helix4D combines efficient cylindrical convolutions with a simplified spatio-temporal transformer architecture operating at low resolution.

\begin{figure}[t]
    \centering
    \input{figures/helixnet_sequences3}
    \caption{\textbf{Coverage from HelixNet.} We split the acquisitions into $12$ training, $2$ validation, and $6$ testing sequences. HelixNet contains diverse scenes in various urban environments from  static or mobile sensors. }
    \label{fig:20sec_HelixNet}
\end{figure}
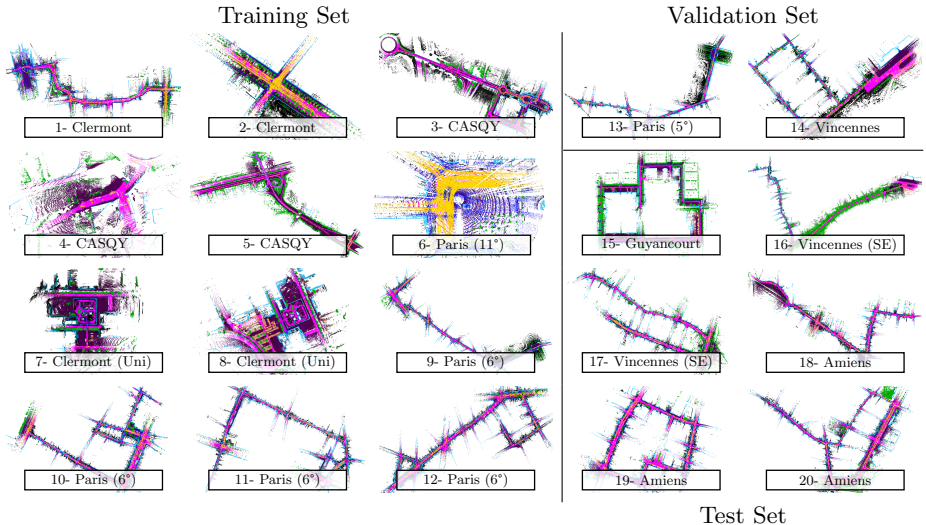

%% file: figures/helixnet_sequences3.tex
\newcommand{\addsequence}[2]{
    {\noindent\begin{tikzpicture}[every node/.style={inner sep=.5pt,outer sep=0}]]
        \node[anchor=south] at (0, 0) {\includegraphics[width=.189\textwidth]{images/HelixNet/sequences/seq#1.png}} ;
        \node[rectangle,draw=black,fill=white, text opacity=1,fill opacity=0.75,inner sep=0.5pt,line width=.5pt, minimum width=.15\textwidth, minimum height=.8em, anchor=south] at (0, .05) {\scalebox{0.6}{#2}};
    \end{tikzpicture}}
}
\begin{tabular*}{\textwidth}{@{}@{\extracolsep{\fill}}*{3}{c}|@{\extracolsep{\fill}}*{2}{c}@{}}
    \multicolumn{3}{c}{Training Set}  & \multicolumn{2}{c}{Validation Set}\\
    \addsequence{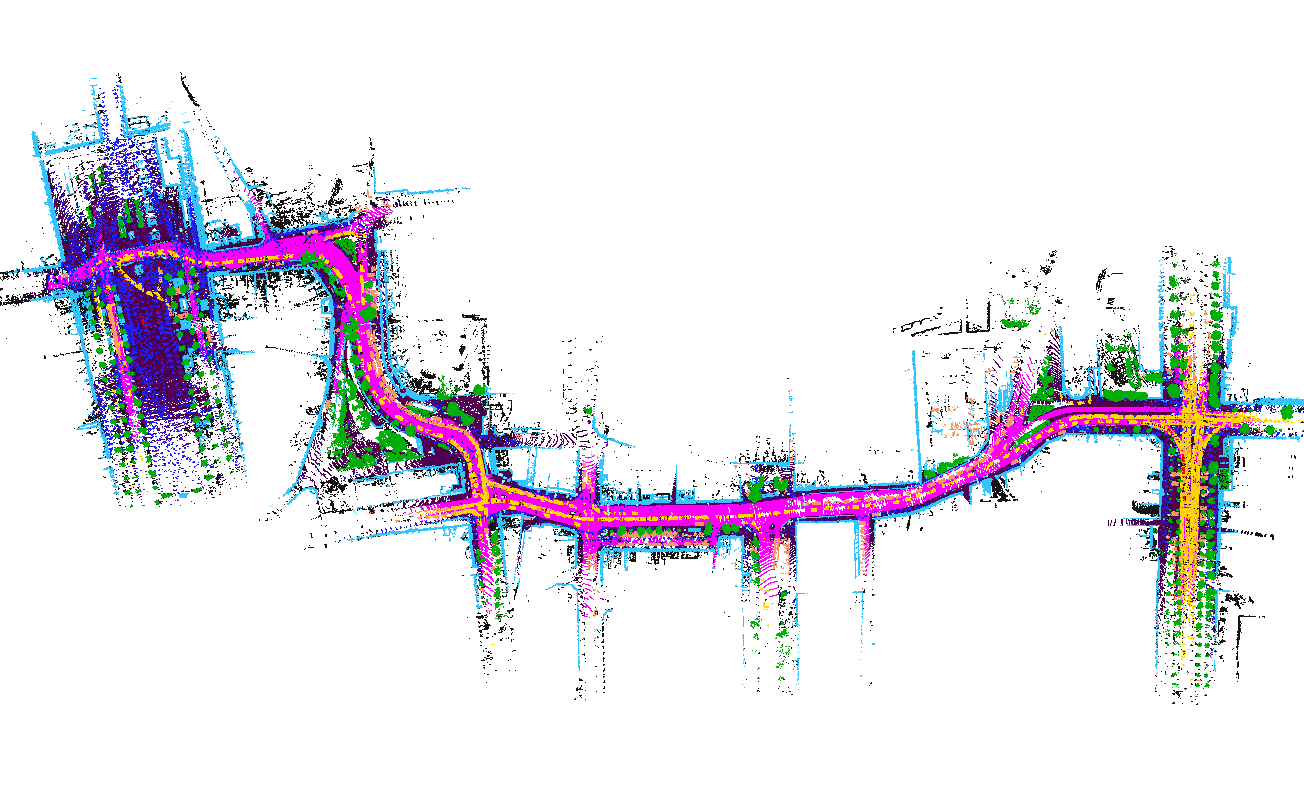}{1- Clermont}& 
    \addsequence{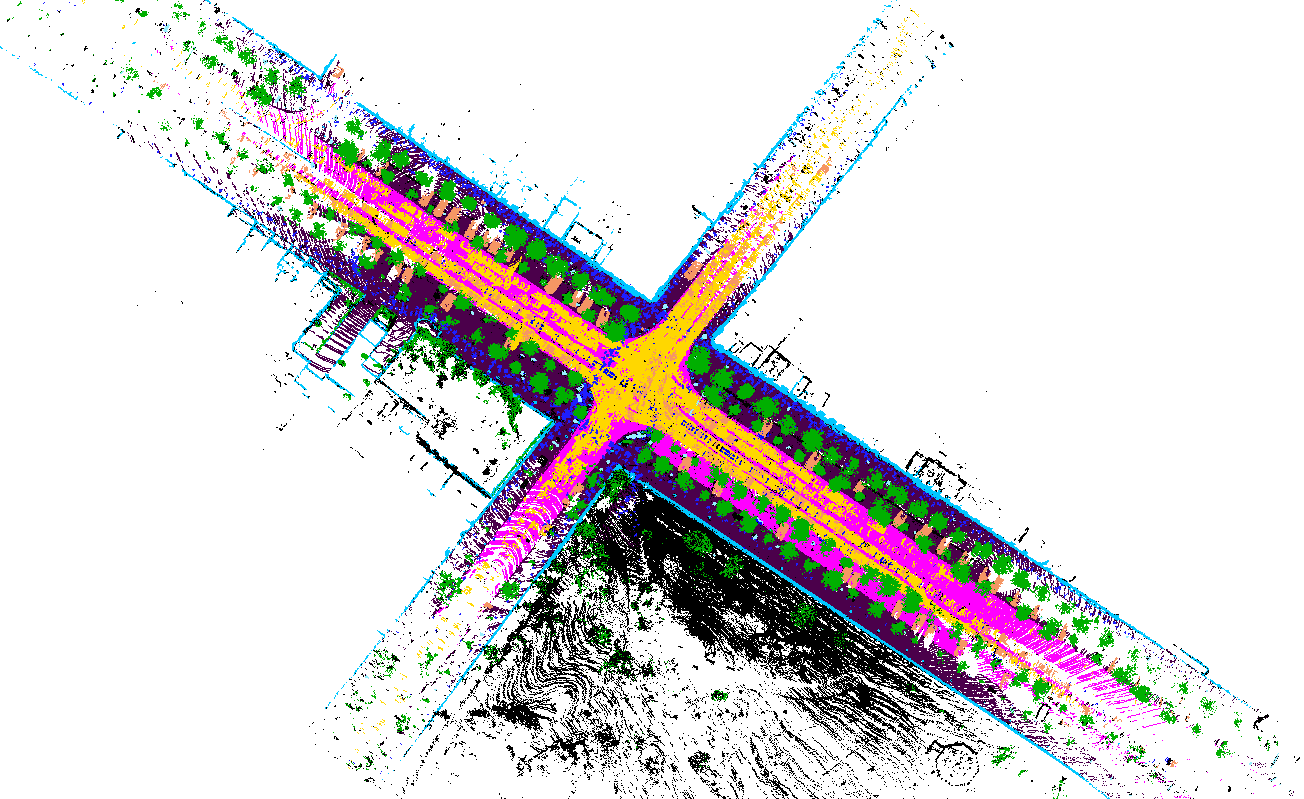}{2- Clermont}& 
    \addsequence{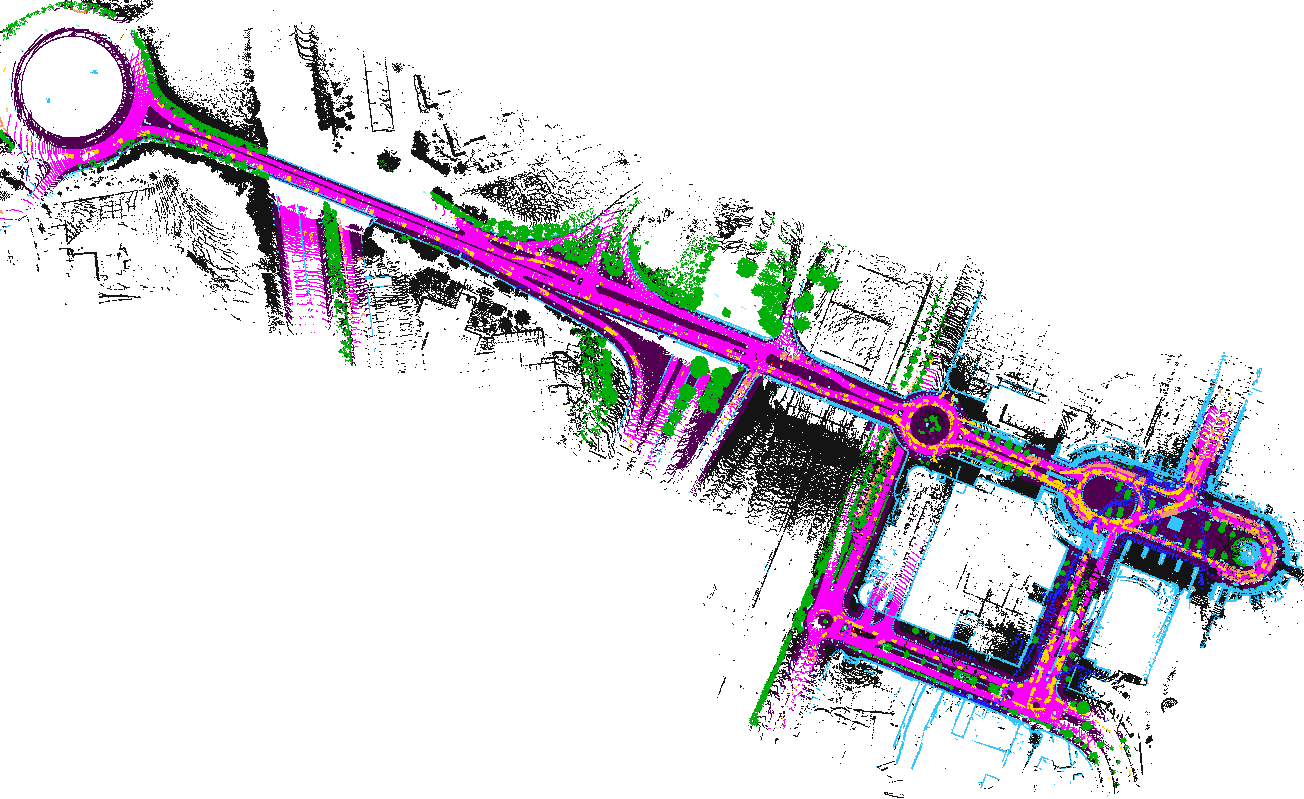}{3- CASQY}&
    \addsequence{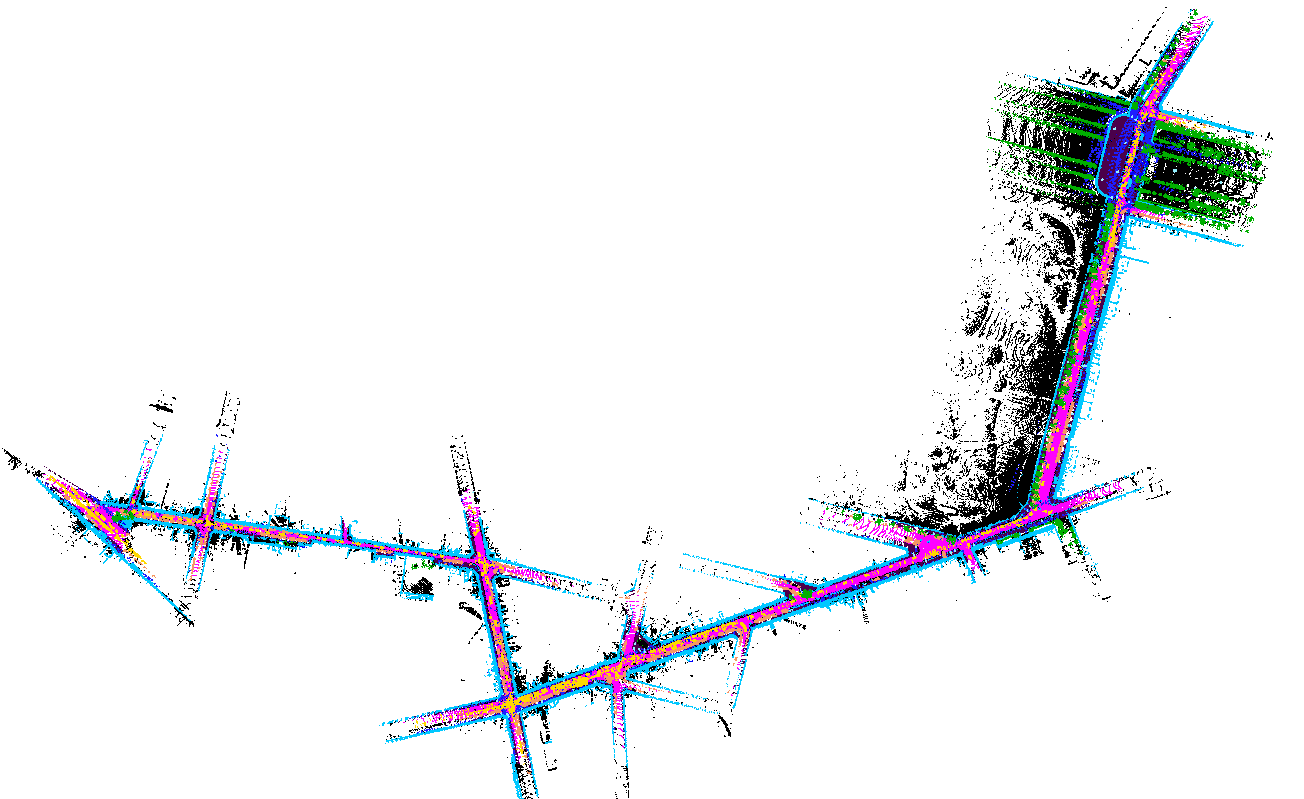}{13- Paris (5°)}&
    \addsequence{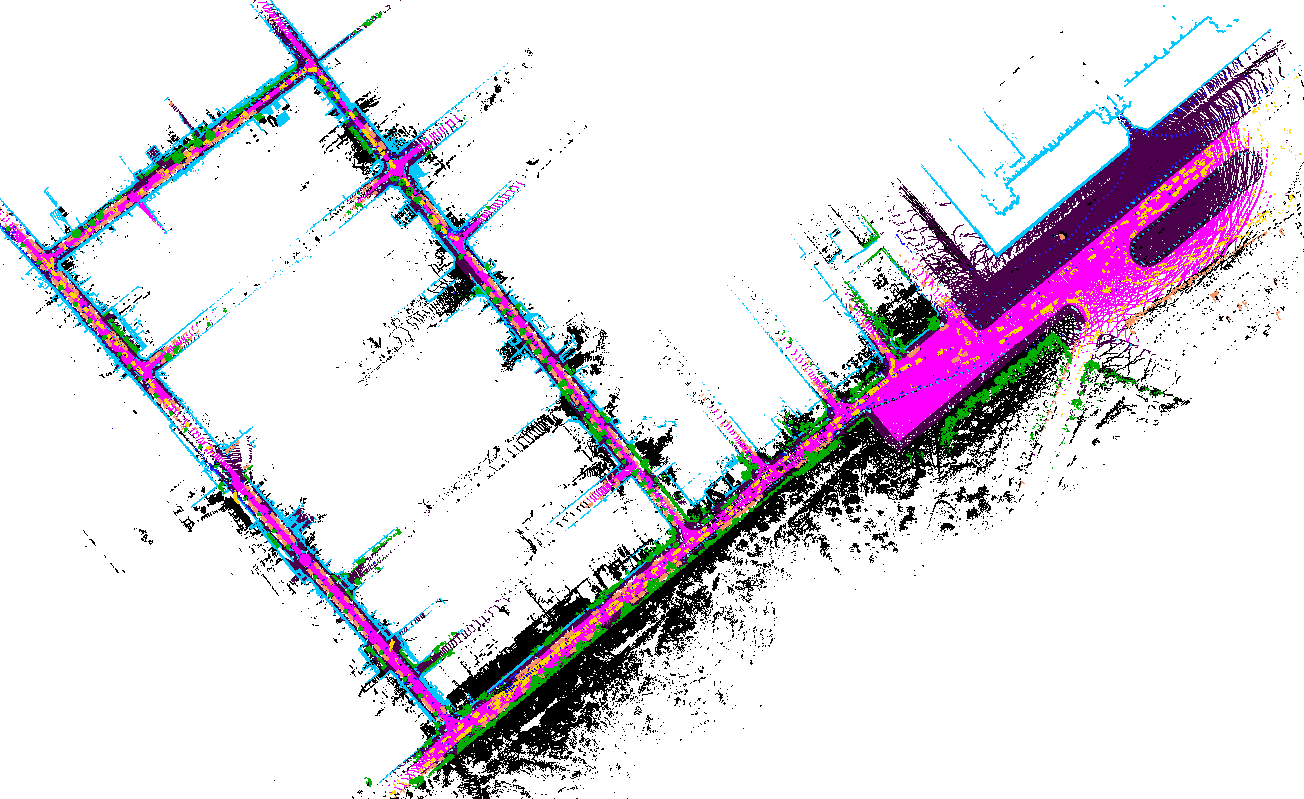}{14- Vincennes}\\\cline{4-5}
    \addsequence{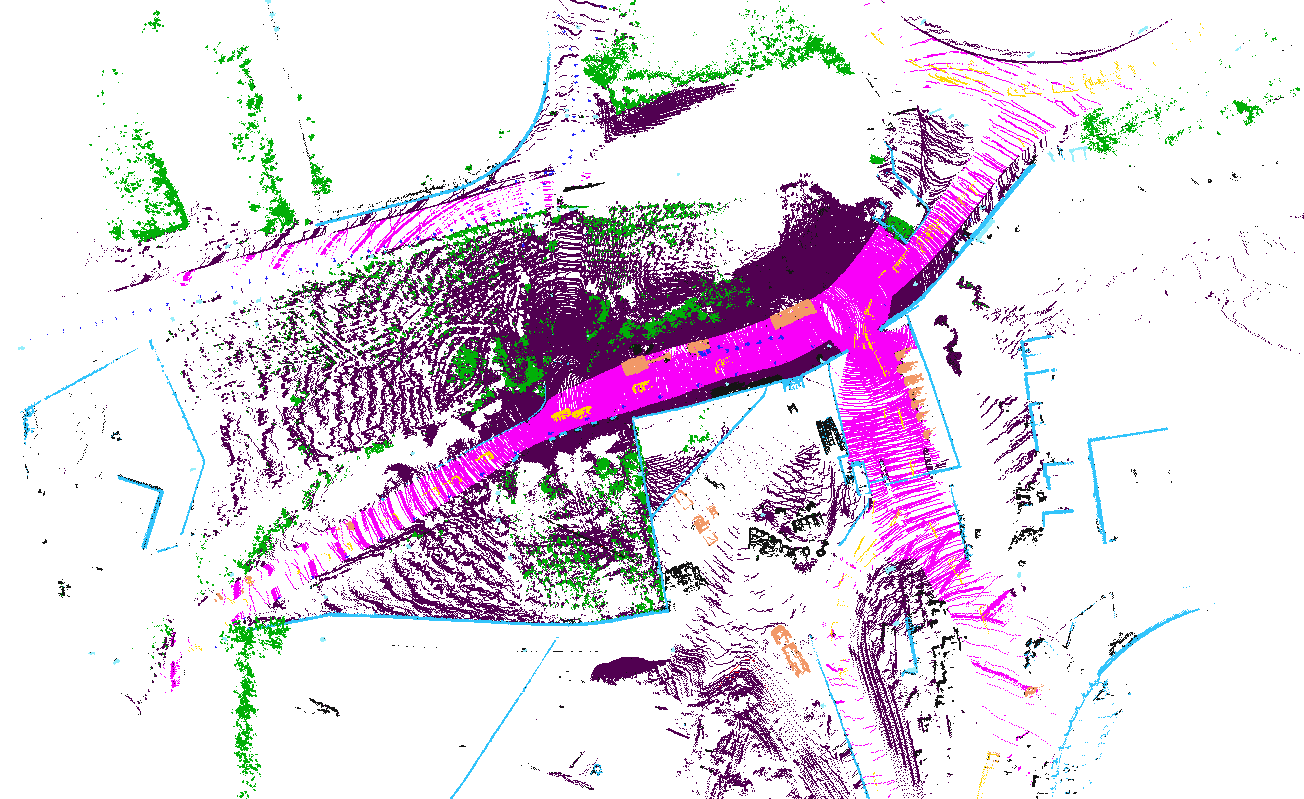}{4- CASQY}& 
    \addsequence{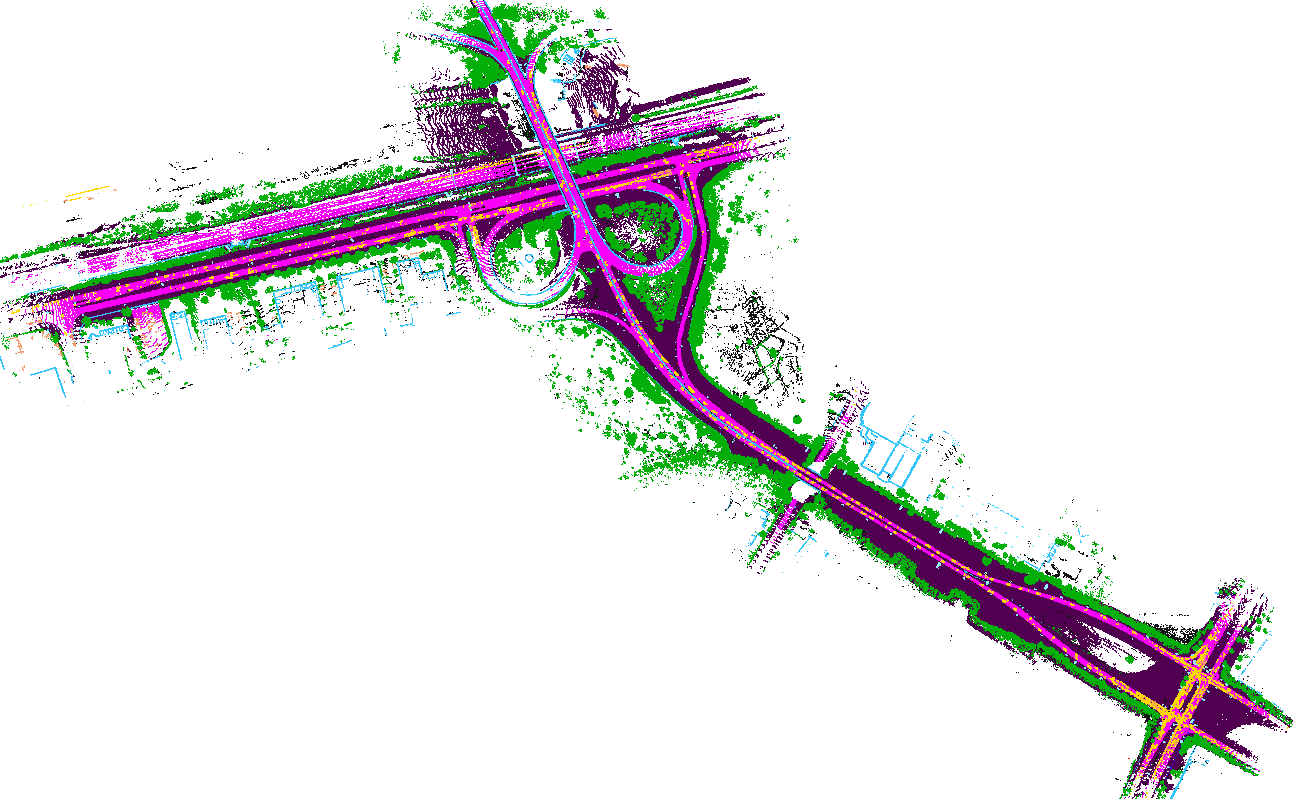}{5- CASQY}& 
    \addsequence{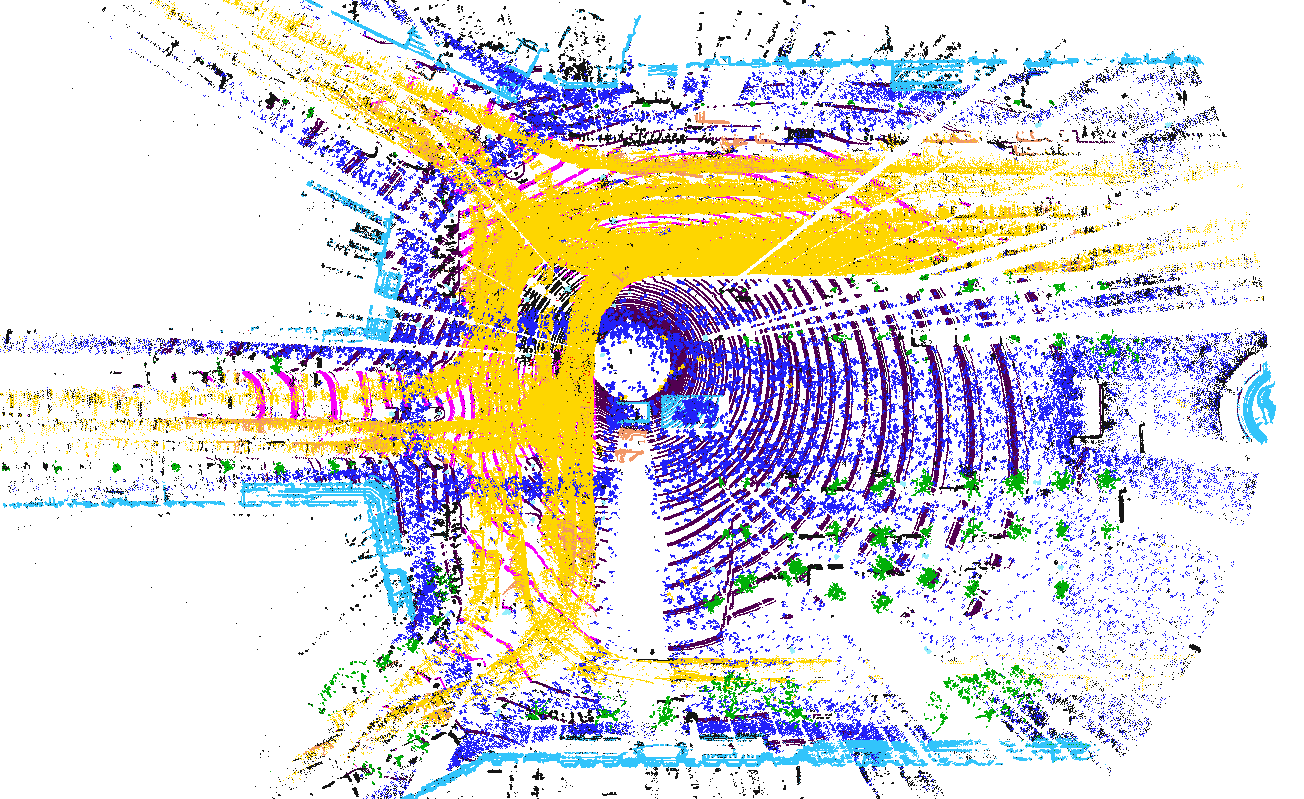}{6- Paris (11°)} &
    \addsequence{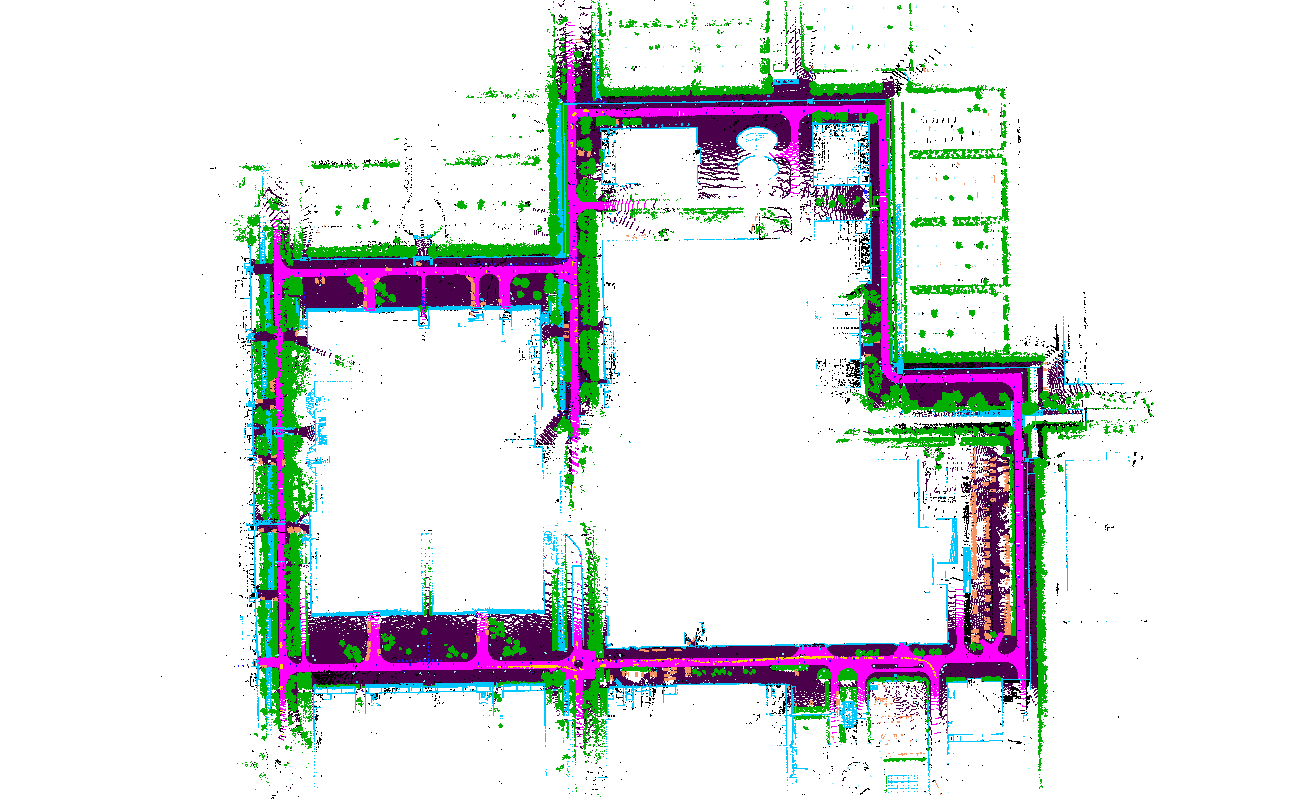}{15- Guyancourt}& 
    \addsequence{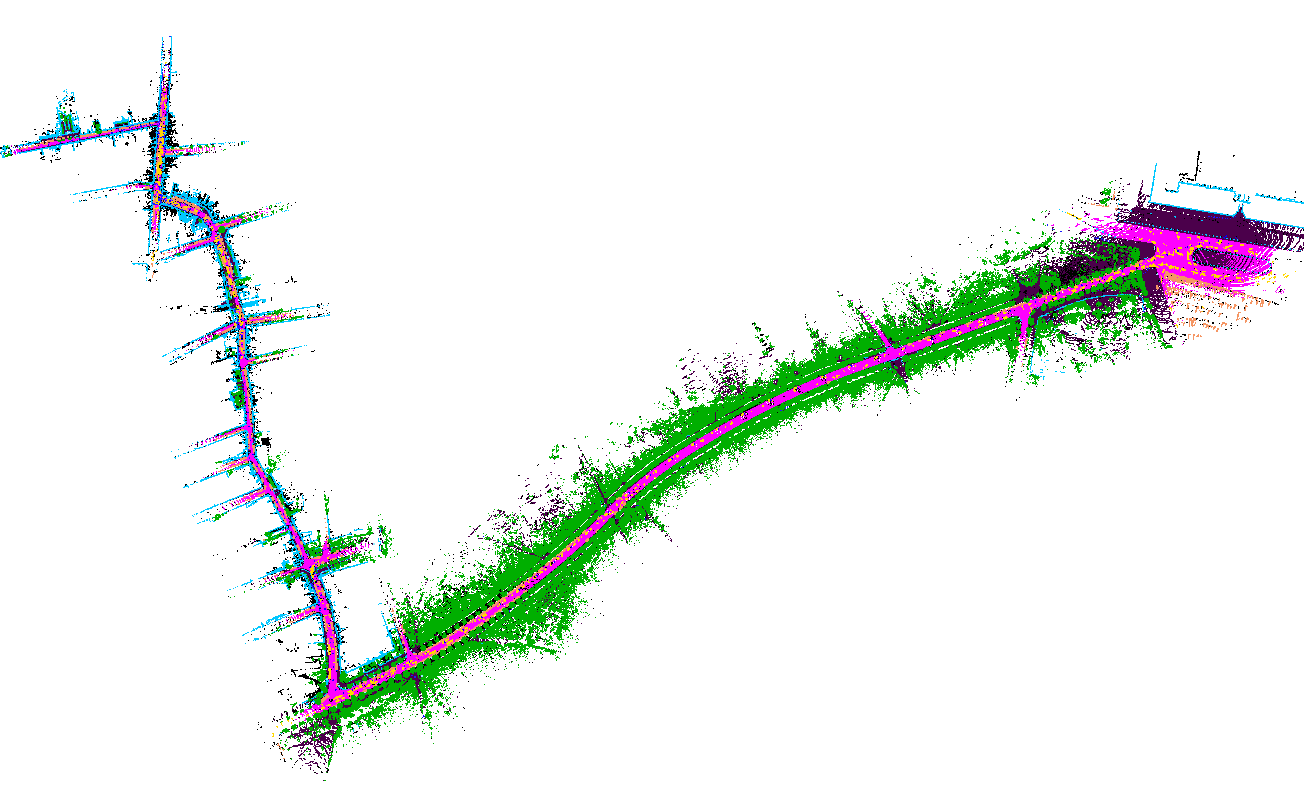}{16- Vincennes (SE)}\\
    \addsequence{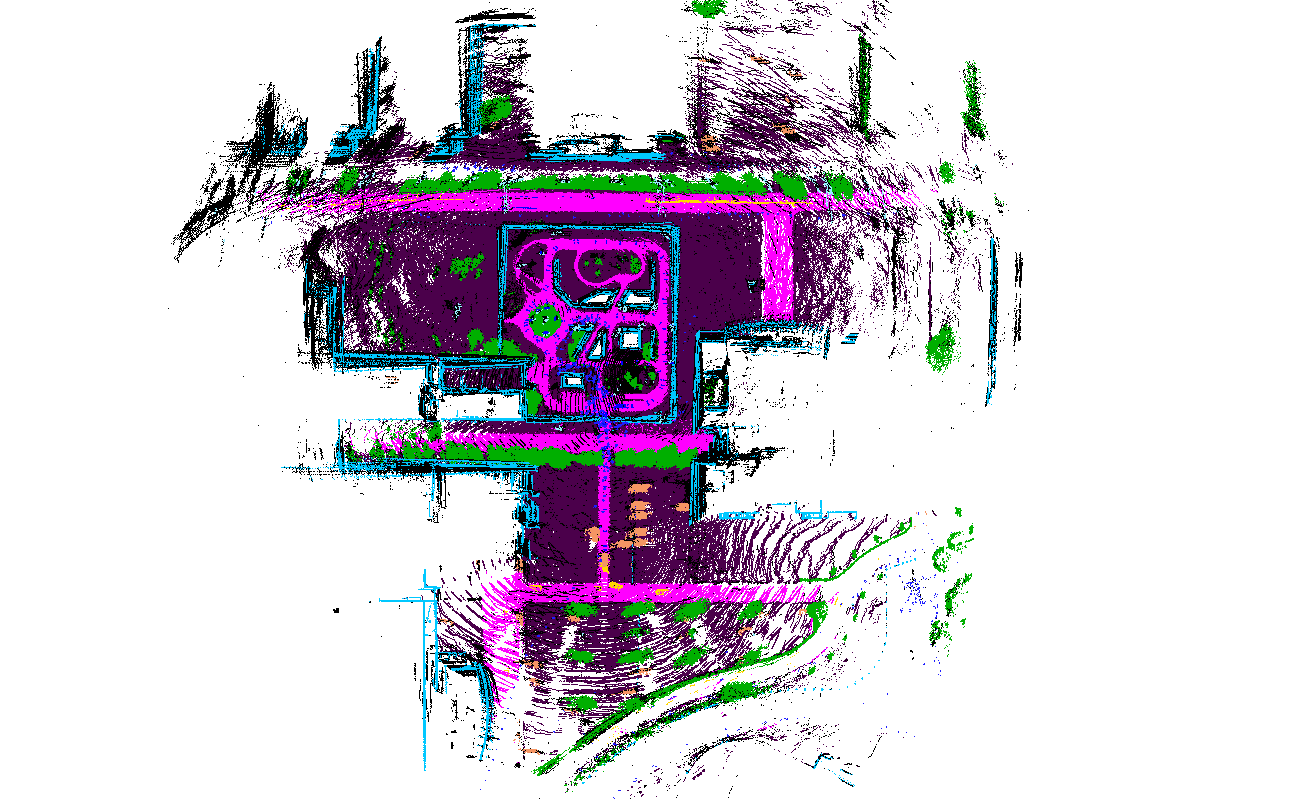}{7- Clermont (Uni)}& 
    \addsequence{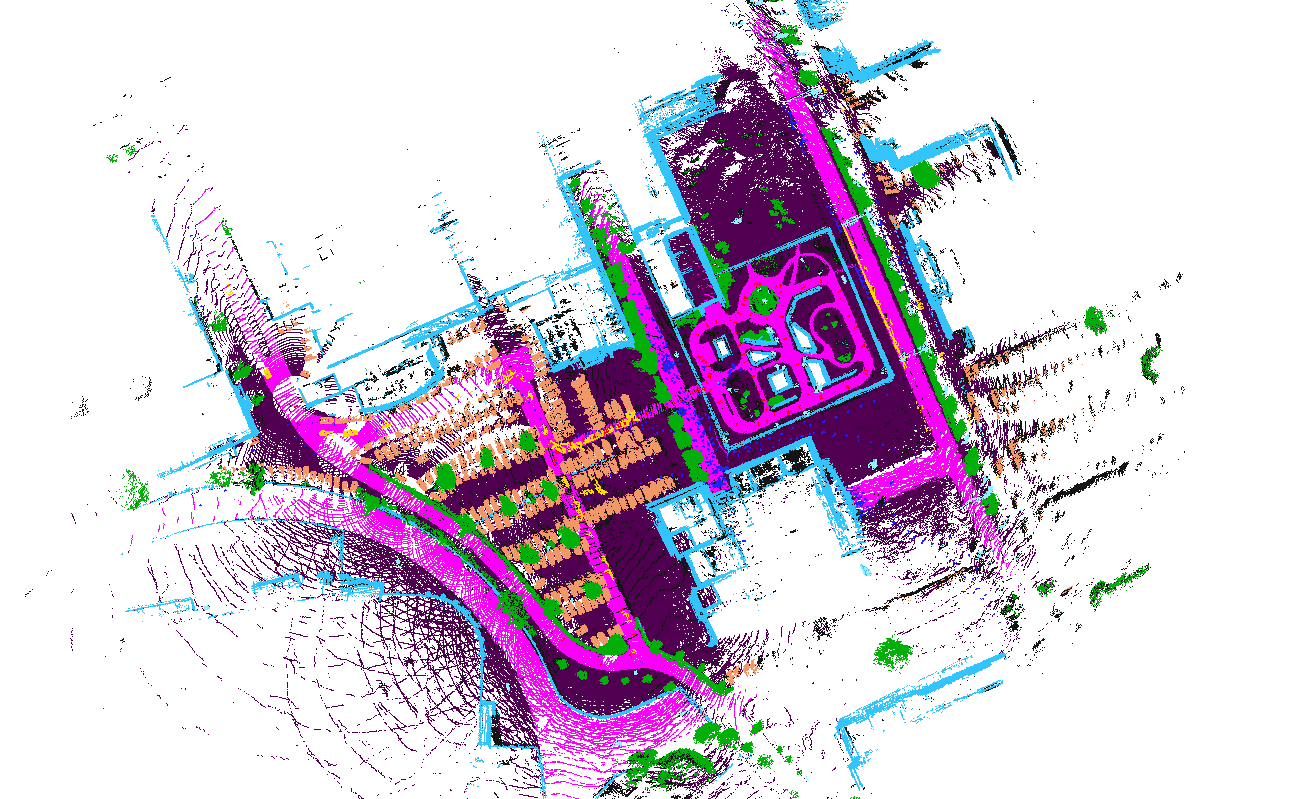}{8- Clermont (Uni)}& 
    \addsequence{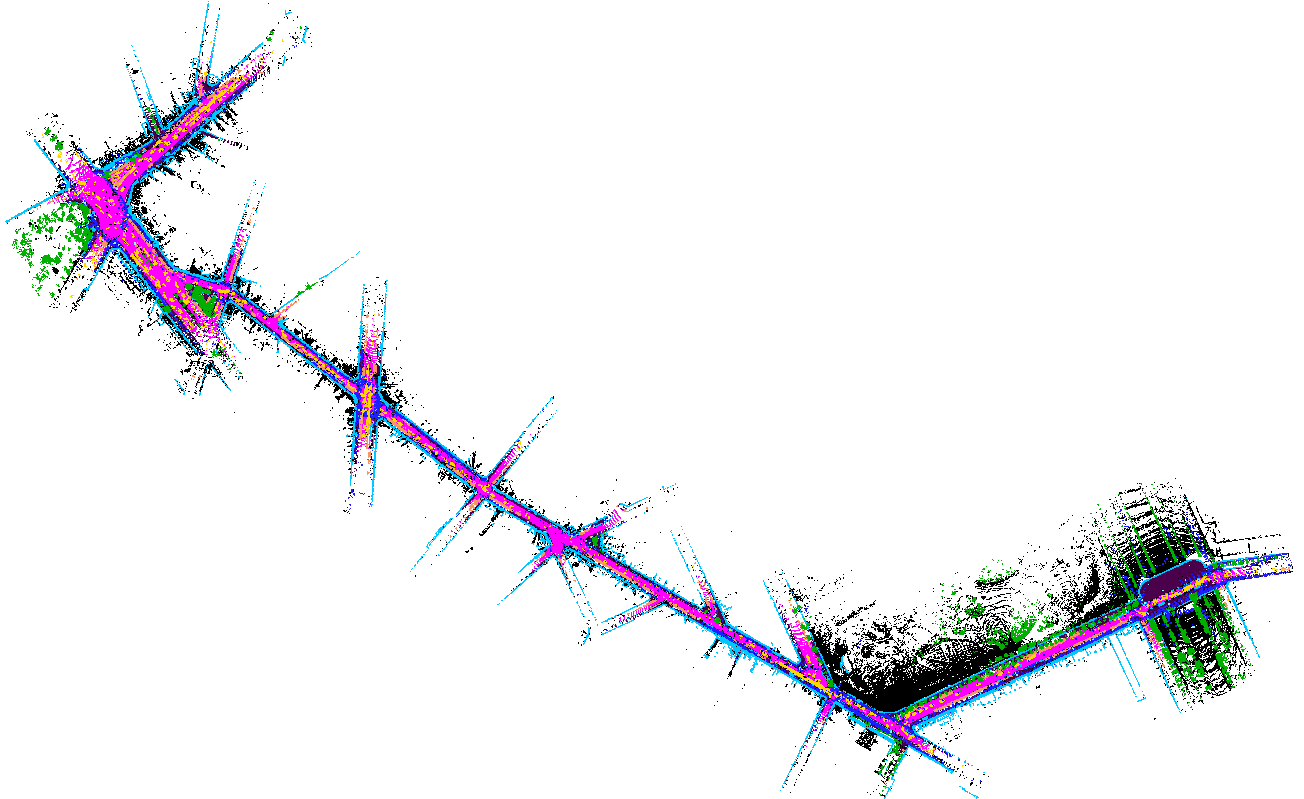}{9- Paris (6°)}& 
    \addsequence{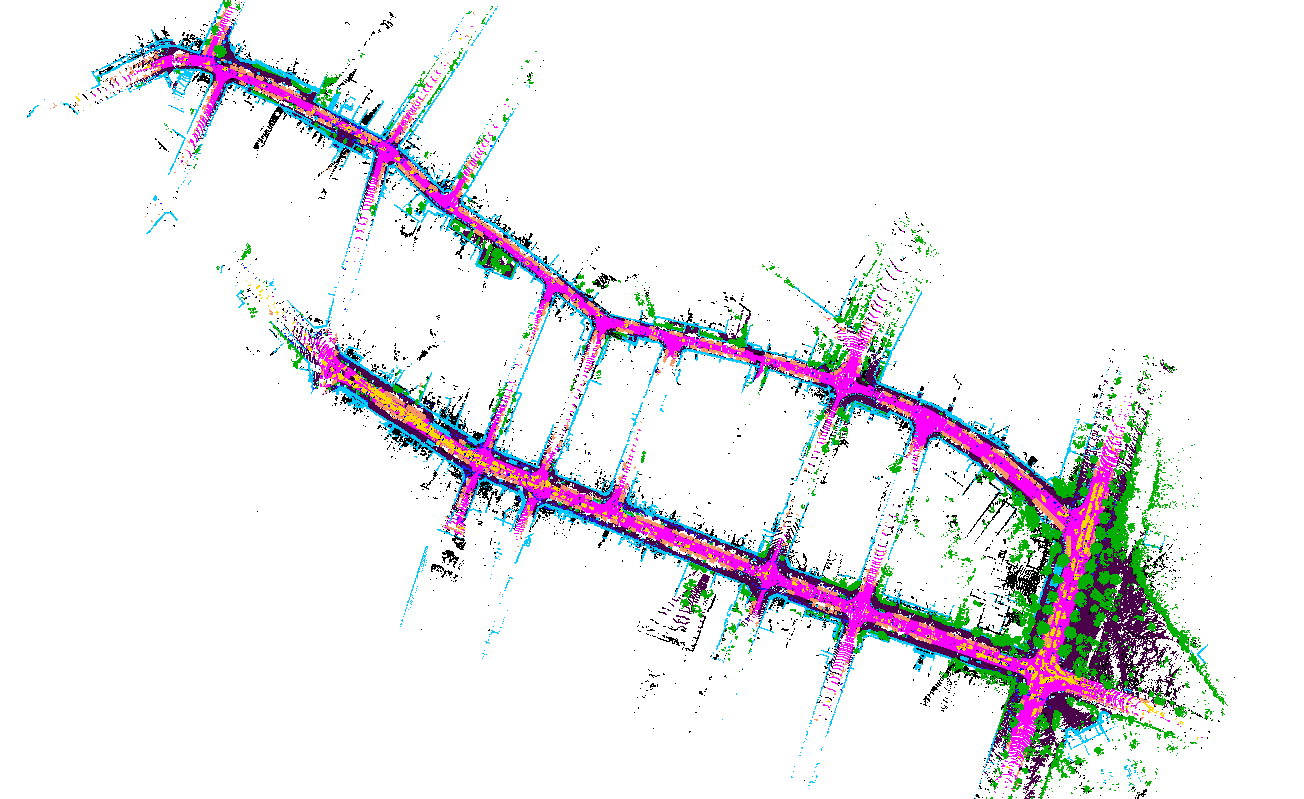}{17- Vincennes (SE)}&
    \addsequence{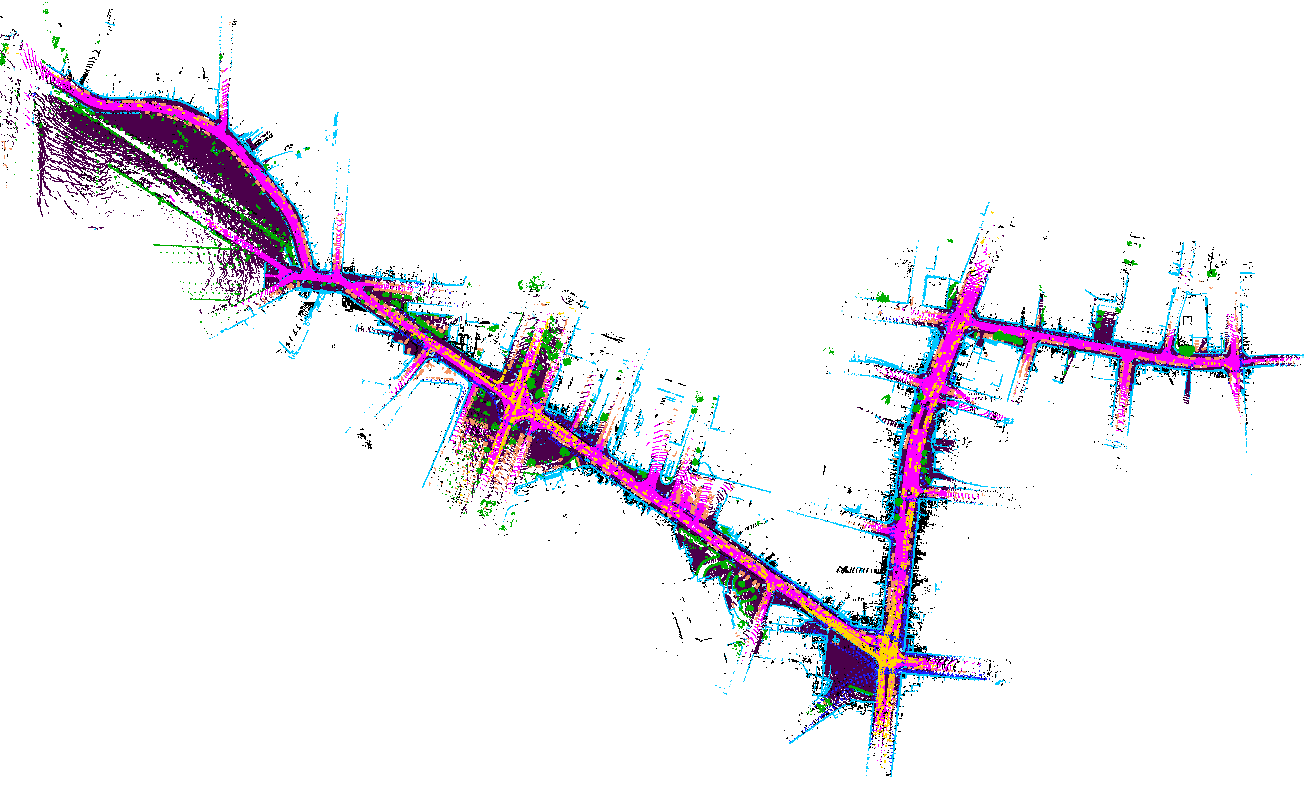}{18- Amiens}\\
    \addsequence{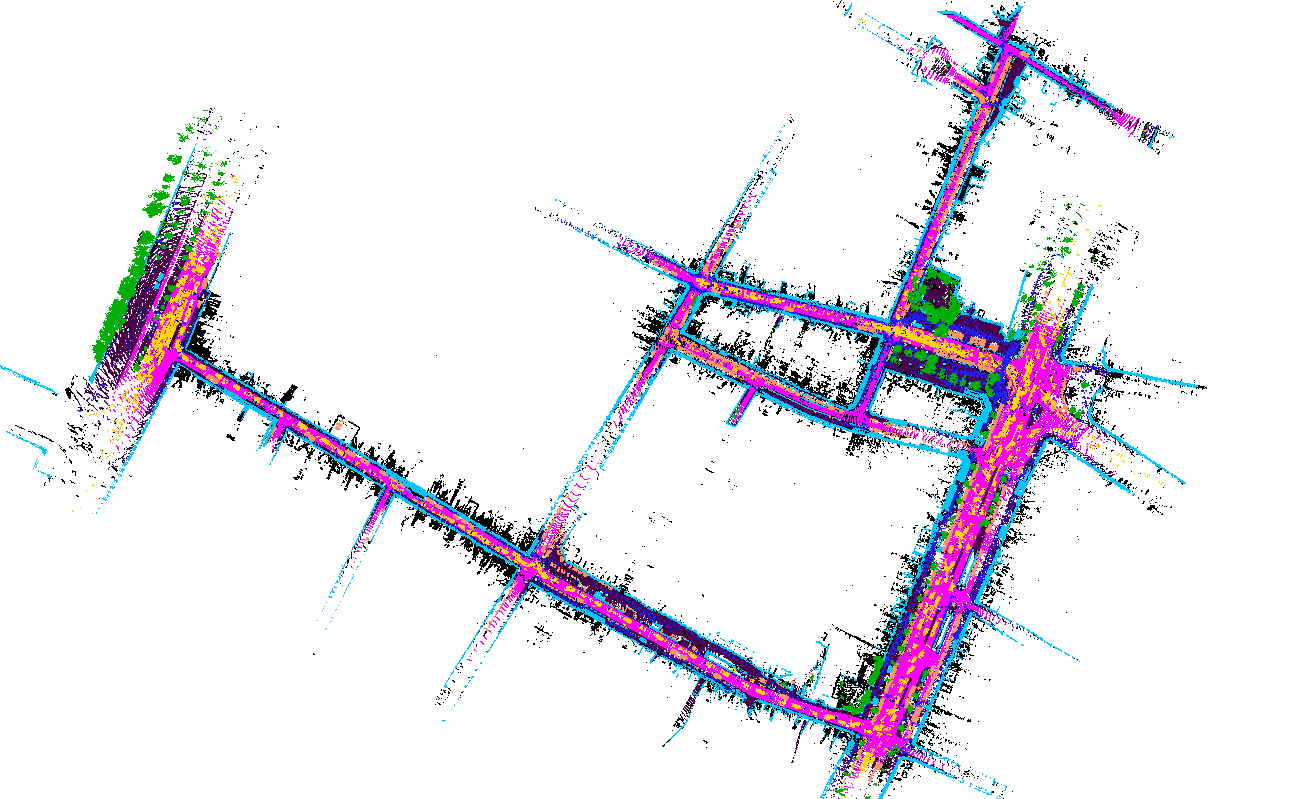}{10- Paris (6°)}& 
    \addsequence{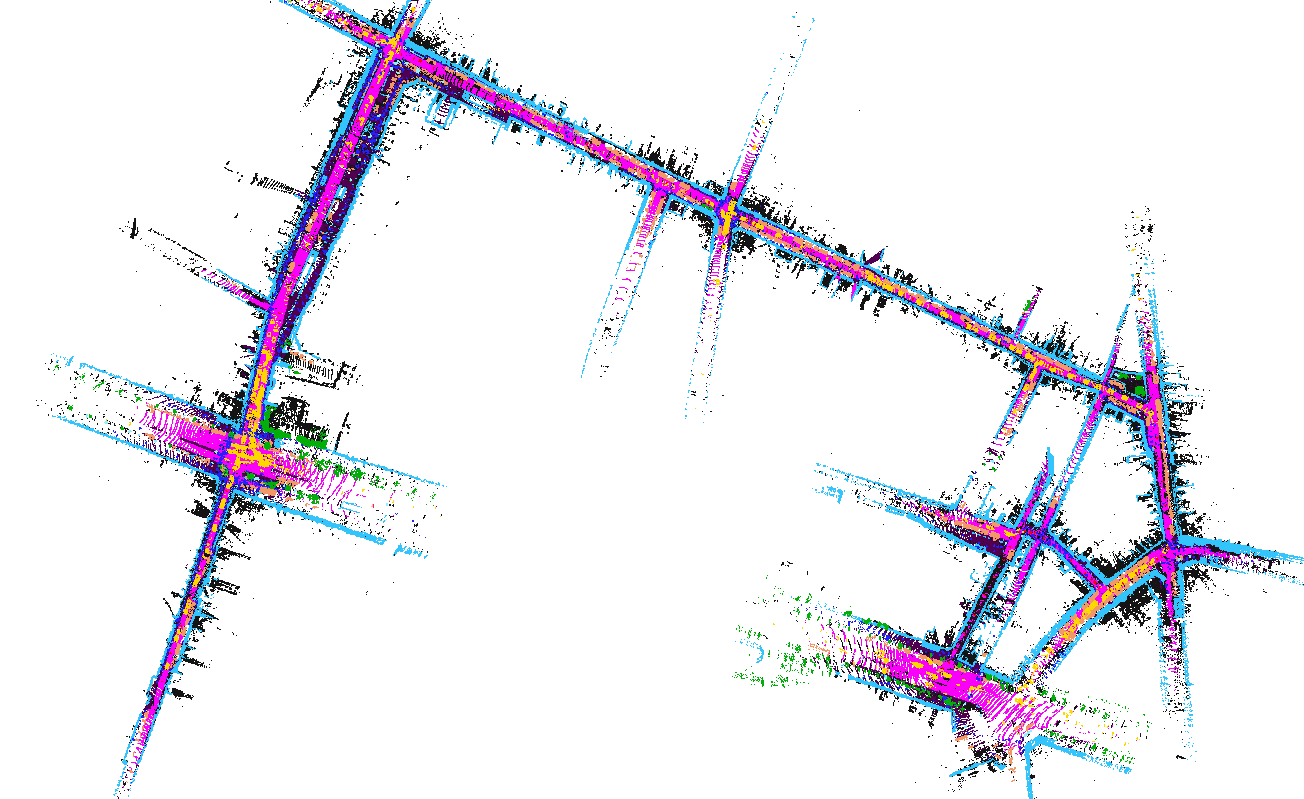}{11- Paris (6°)}& 
    \addsequence{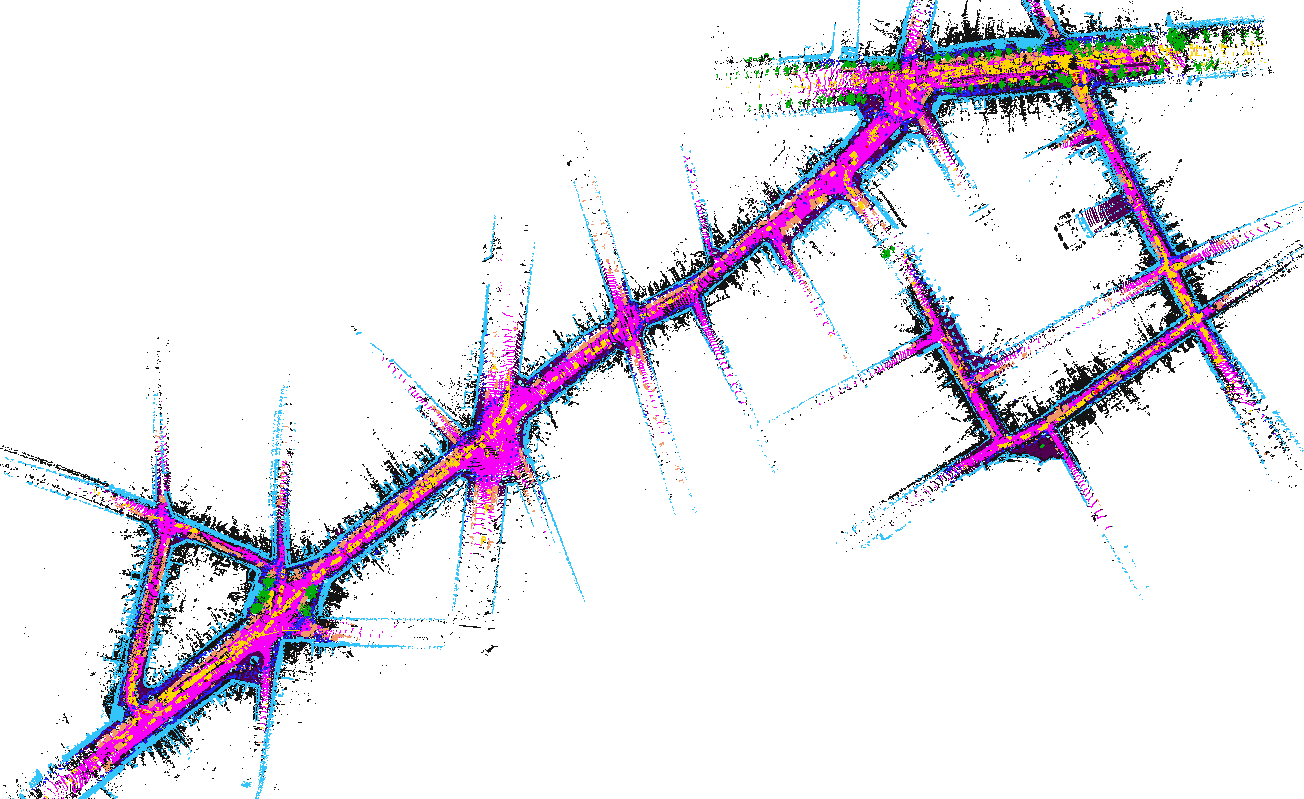}{12- Paris (6°)}&
    \addsequence{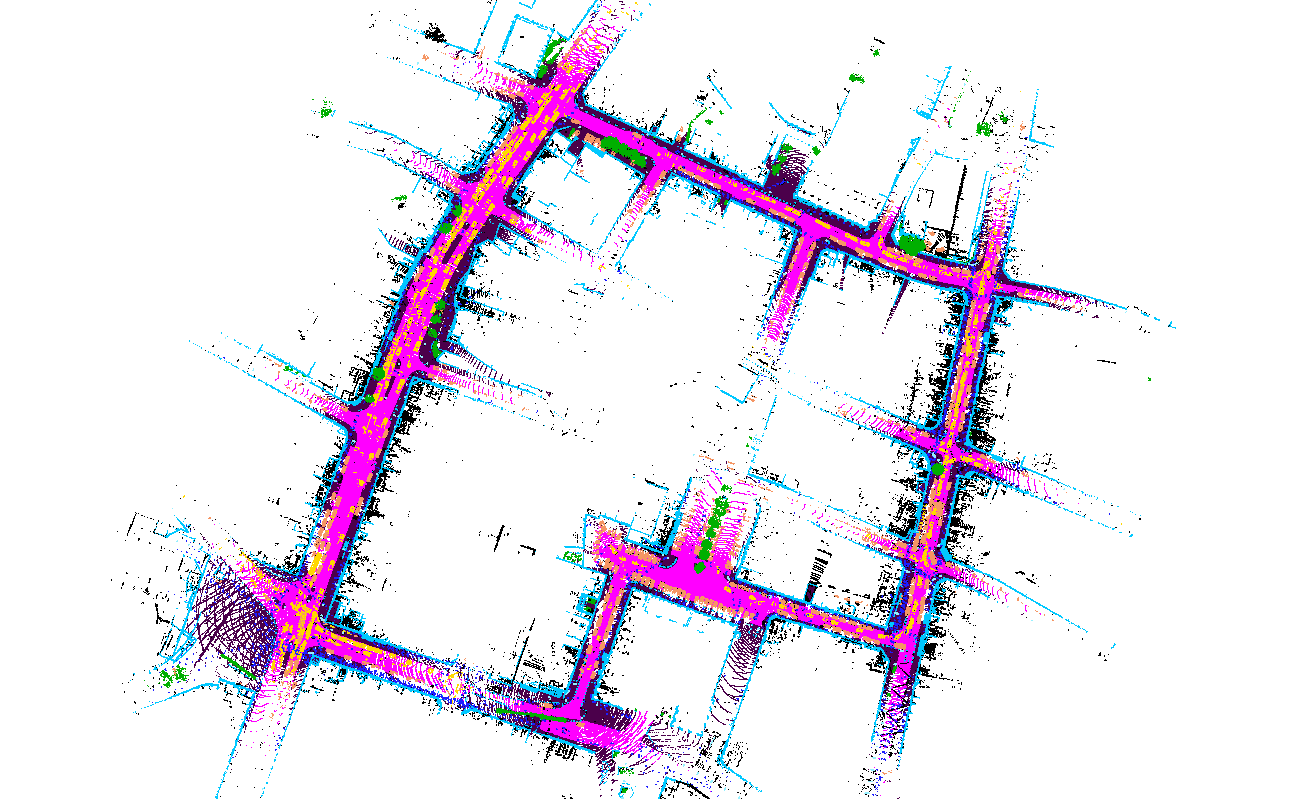}{19- Amiens}& 
    \addsequence{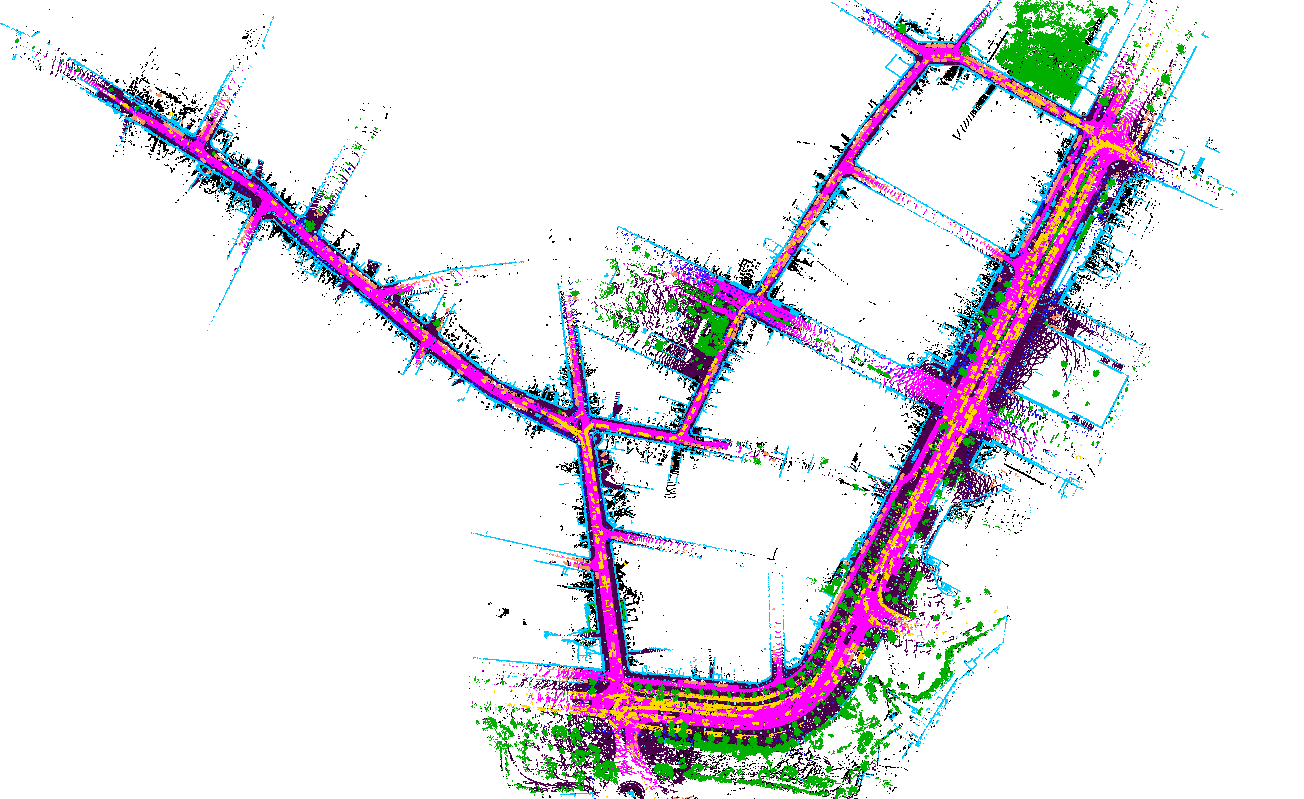}{20- Amiens}\\
    \multicolumn{3}{c}{} & \multicolumn{2}{c}{Test Set}\\
\end{tabular*}

%% file: text/FINAL/30_helixnet.tex
\section{HelixNet: A Dataset for Online LiDAR Segmentation }\label{sec:HelixNet}

\begin{figure}[t]
    \centering
    \input{figures/helixnet_zooms}
    \caption{\textbf{Extracts from HelixNet.}~Our proposed dataset contains various urban scenes from motorway to pedestrian plazas and historical centers. In the first row, we represent extracts of $15$ to $30$s of acquisition colored according to the point release time. In the second row, we represent the point semantic labels.}
    \label{fig:zoom_HelixNet}
\end{figure}
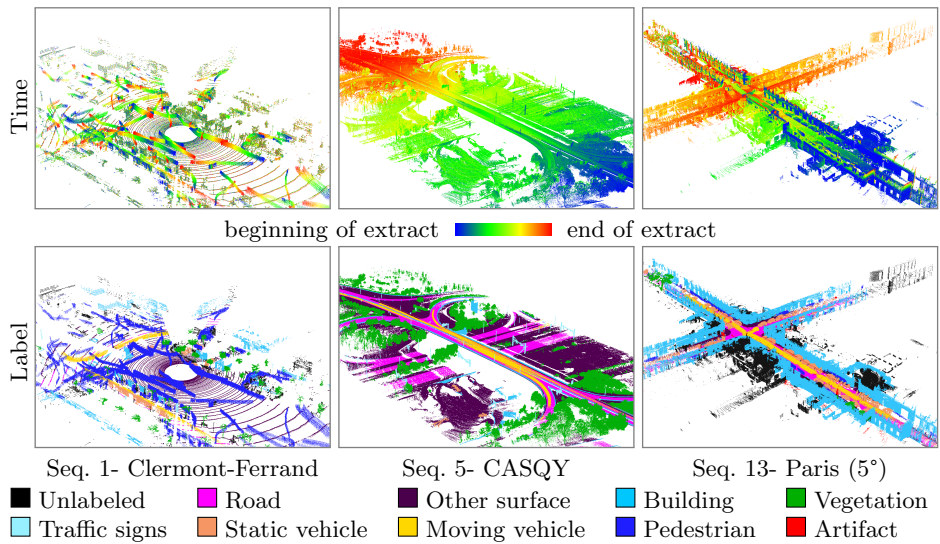

We introduce HelixNet, a new large-scale and open-access LiDAR dataset intended for the evaluation of real-time semantic segmentation algorithms. In contrast to other large-scale datasets, HelixNet includes fine-grained data on sensor rotation and position, as well as point release time. 

\paragraph{General Characteristics.}
As seen in \figref{fig:20sec_HelixNet}, HelixNet contains $20$ sequences of 3D points, each corresponding to $6$ to $7$ minutes of continuous acquisition, for a total of $129$ minutes. Scanning was performed by an HDL-64E Velodyne rotating LiDAR~\cite{velodyne} mounted on a 
mobile platform~\cite{stereopolis}. 
As shown in \figref{fig:zoom_HelixNet}, HelixNet covers multiple cities and a wide variety of environments such as a university campus, dense historical centers, and a highway interchange.
With a total of $10$ billion points across $78\,800$ frames and $8.85$ billion individual labels, HelixNet is the largest densely annotated open-access rotating LiDAR dataset by a factor of $1.7$ as shown in \tabref{tab:helixnetvsothers}. 
HelixNet follows the file format of SemanticKITTI~\cite{behley2019iccv}, 
allowing researchers to evaluate existing code with minimal effort.

We use a $9$-classes nomenclature: \emph{road}~($16.4\%$ of all points), \emph{other surface}~($22.0\%$), \emph{building}~($31.3\%$), \emph{vegetation}~($8.5\%$), \emph{traffic signs}~($1.6\%$), \emph{static vehicle}~($4.9\%$), \emph{moving vehicle}~($2.1\%$), \emph{pedestrian}~($0.9\%$), and \emph{acquisition artifact}~($0.05\%$). 
Points without labels correspond to either \emph{un-annotated}~($6.2\%$) parts of the clouds due to their ambiguity, or point without echos ($6.1\%$).
Compared to fine-grained classes such as the ones used by SemanticKITTI \cite{behley2019iccv} or Paris-Lille3D's \cite{roynard2017parislille3d}, our focused nomenclature limits class imbalance and makes macro-averaged metrics more stable.

Each point is associated with the $9$ following values: (1-3) Cartesian coordinates in a fixed frame of a reference, (4-6) cylindrical coordinate relative to the sensor at the time of acquisition, (7) intensity, (8) fiber index, and (9) packet output time. As detailed in the next paragraph, the last two features are not typically available in large-scale datasets and cannot be inferred. 

\paragraph{Sensor-Based Timing and Grouping.}
A rotating LiDAR consists of a set of lasers---or fibers---arranged on a rotating sensor head. The lasers send periodic pulses of light whose return times give the position of the impact points relative to the sensor. In the context of autonomous driving, these sensors are typically deployed on a moving platform and capture 3D points with centimetric accuracy.
The sensor releases the data stream as a discrete temporal sequence of \emph{packets} of 3D points. For an HDL-64E LiDAR, each packet contains $6\times64$ points,  corresponding to around $1^\circ$ rotation of the sensor. To represent the real-time operational setting of autonomous driving, we associate with each point the timestamp of its \emph{packet output event}, \ie the instant the packet is available and not the acquisition time of the point. The latency between the acquisition of the first point and the complete transfer of its packet is $278\mu$s. Although small compared to acquisition and inference times, this more rigorous timing constitutes a step towards a more realistic evaluation setting of segmentation algorithms of LiDAR sequences. 

\begin{figure}[t]
    \centering
    \begin{subfigure}[b]{0.495\textwidth}
            \includegraphics[width=\textwidth, clip, trim=0 3pt 10pt 0]{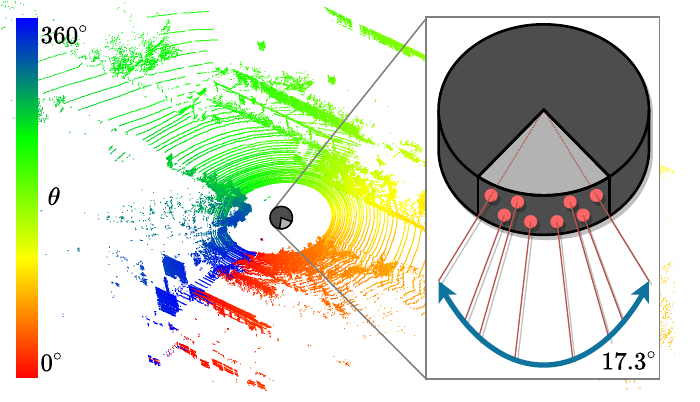}
        \subcaption{Rotation of the sensor head}
        \label{fig:sub:theta}
    \end{subfigure}
    \hfill
    \begin{subfigure}[b]{0.495\textwidth}
            \includegraphics[width=\textwidth, clip, trim=0 3pt 0 0]{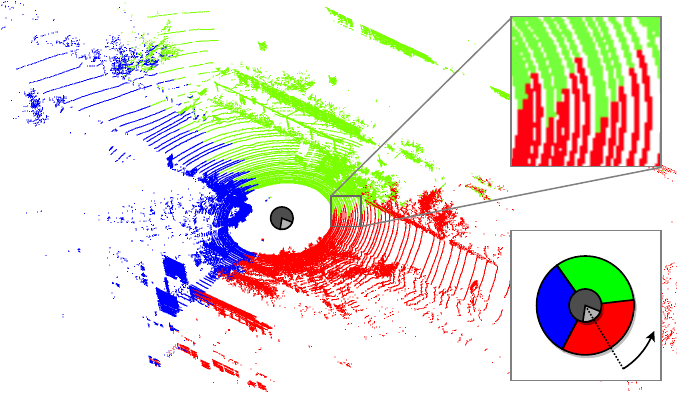}
        \subcaption{Slices covering $120^\circ$}
        \label{fig:sub:slices}
    \end{subfigure}
    \caption{\textbf{Sensor Acquisition Geometry.}~
    We represent in \Subref{fig:sub:theta} the acquisition of a rotating sensor, which is split into \sfrac{1}{3} turn slices in \Subref{fig:sub:slices}. 
    As the Laser emitters position forms an angle of over $17.3^\circ$ around the sensor head, taking slices with respect to the sensor rotation $\theta$ results in a jagged profile.
    }
    \label{fig:jaggedslices}
\end{figure}

On top of its absolute position, we associate with each individual point its cylindrical coordinates relative to the position of the sensor at the exact time of its acquisition. This differs from other datasets such as SemanticKITTI~\cite{behley2019iccv}, which gives the relative position of all points but the absolute position of the sensor only once per frame. 
While sensor movement can be interpolated, the vehicle trajectory might not be linear and the sensor head rotates. For comparison, at $50$km/h, the sensor moves more than $1.4$m during each rotation.

LiDAR sequences are typically split into frames containing points that cover a $360^\circ$ degree arc around the sensor. However, the acquisition geometry makes this grouping artificial.
Indeed, the fibers (\ie the individual lasers) do not all face the same direction: they are arranged around the sensor's heads at different angles, with a range of more than $17.3^\circ$. This means that the points within a packet are not vertically aligned but present a jagged profile as seen in \figref{fig:jaggedslices}. In order to obtain frames with \emph{straight edges} such as those of SemanticKITTI~\cite{behley2019iccv}, we would have to consider an acquisition over a sensor rotation of $377^\circ$, adding a further $5$ms of latency.
Contrary to other datasets, HelixNet contains the index of the emitter of each point and organizes the points with respect to the angle of the sensor itself
This allows us to easily build frames or frame portions that are directly consistent with the rotation \emph{of the sensor head itself}. This is important for measuring the real latency of segmentation methods and, as described in the next section, contributes to the efficiency of our proposed network.

%% file: figures/helixnet_zooms.tex
\begin{tabular*}{\textwidth}{@{}@{\extracolsep{\fill}}*{4}{c}@{}}
    \rotatebox{90}{~~~~~~~~~Time} &
    \includegraphicszoomhelixnet{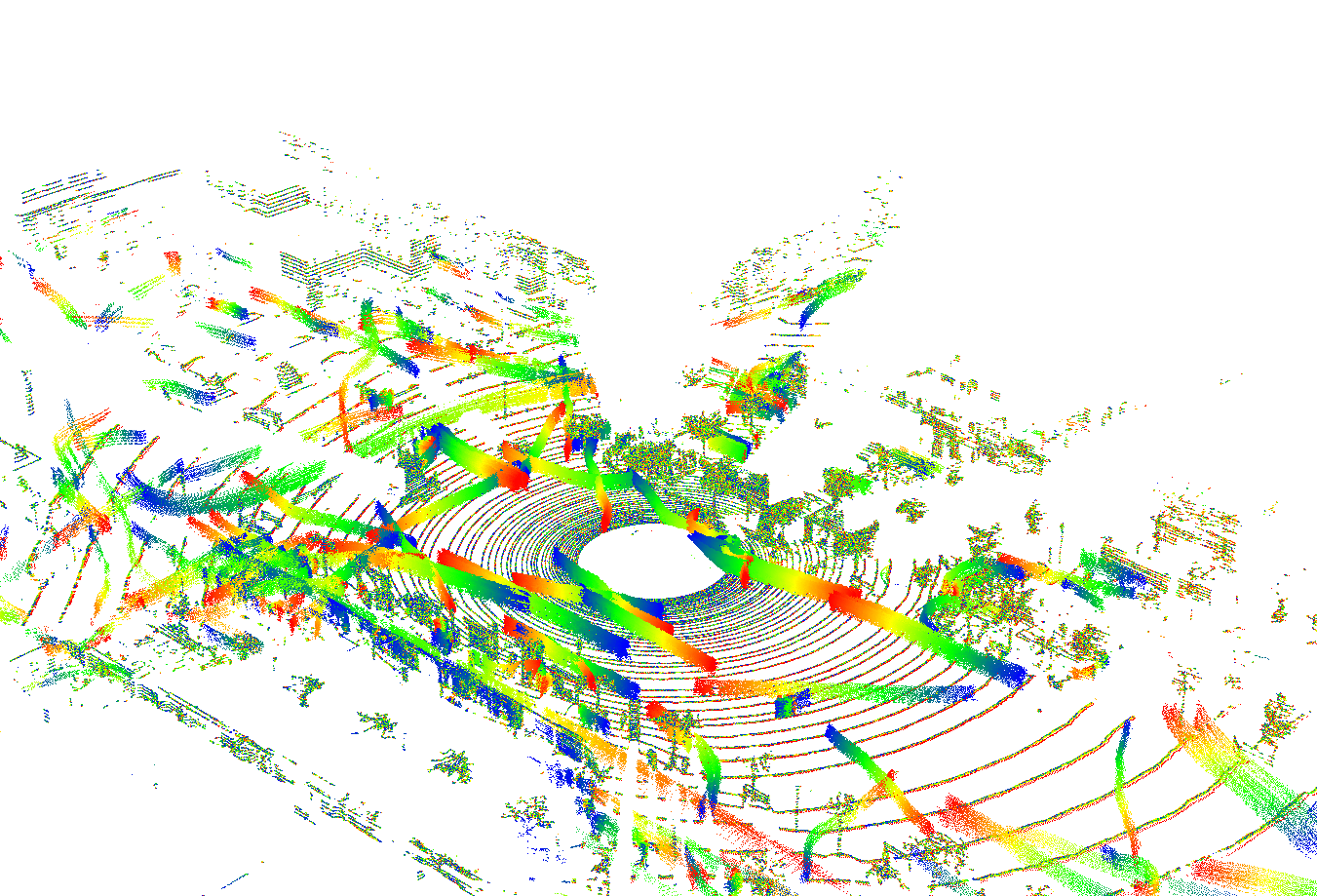} &
    \includegraphicszoomhelixnet{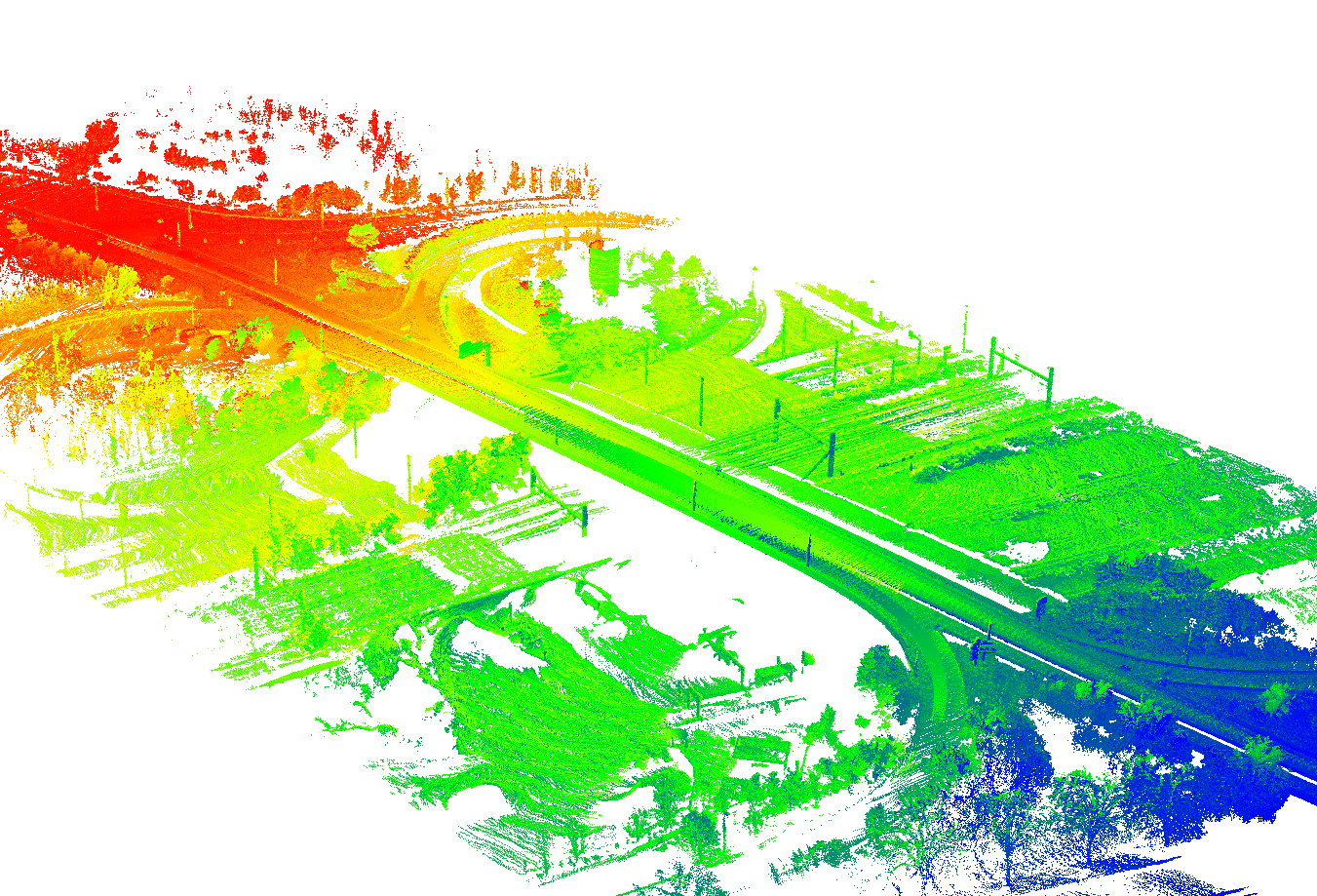} &  
    \includegraphicszoomhelixnet{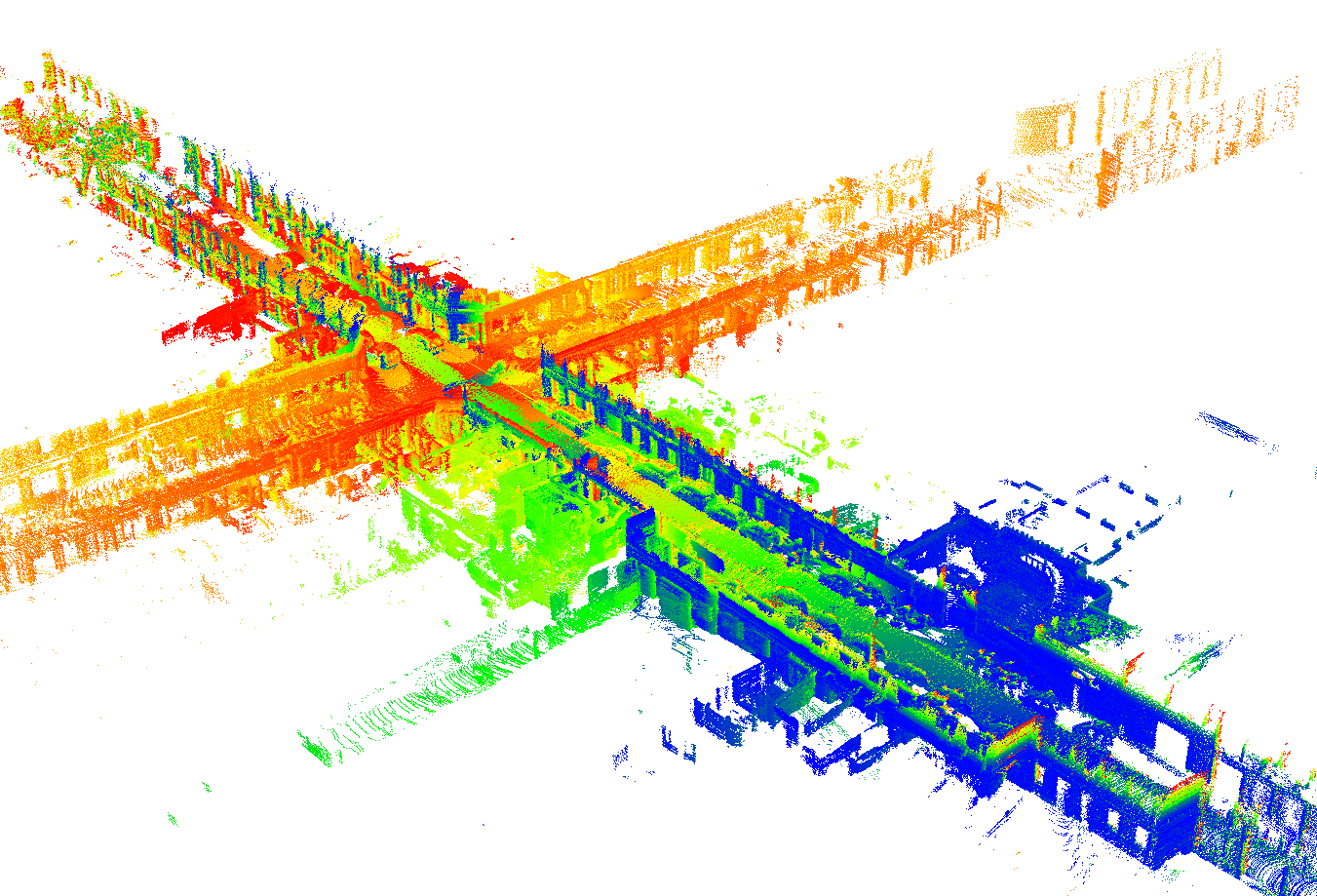} 
    \\
    \multicolumn{4}{c}
    {
    \begin{tabular}{rcl}
        \hfill
        beginning of extract
        & 
        \intextcolorscale{myshade}
        & 
        end of extract
        \hfill
    \end{tabular}
    }\\
    \rotatebox{90}{~~~~~~~~Label} &
    \includegraphicszoomhelixnet{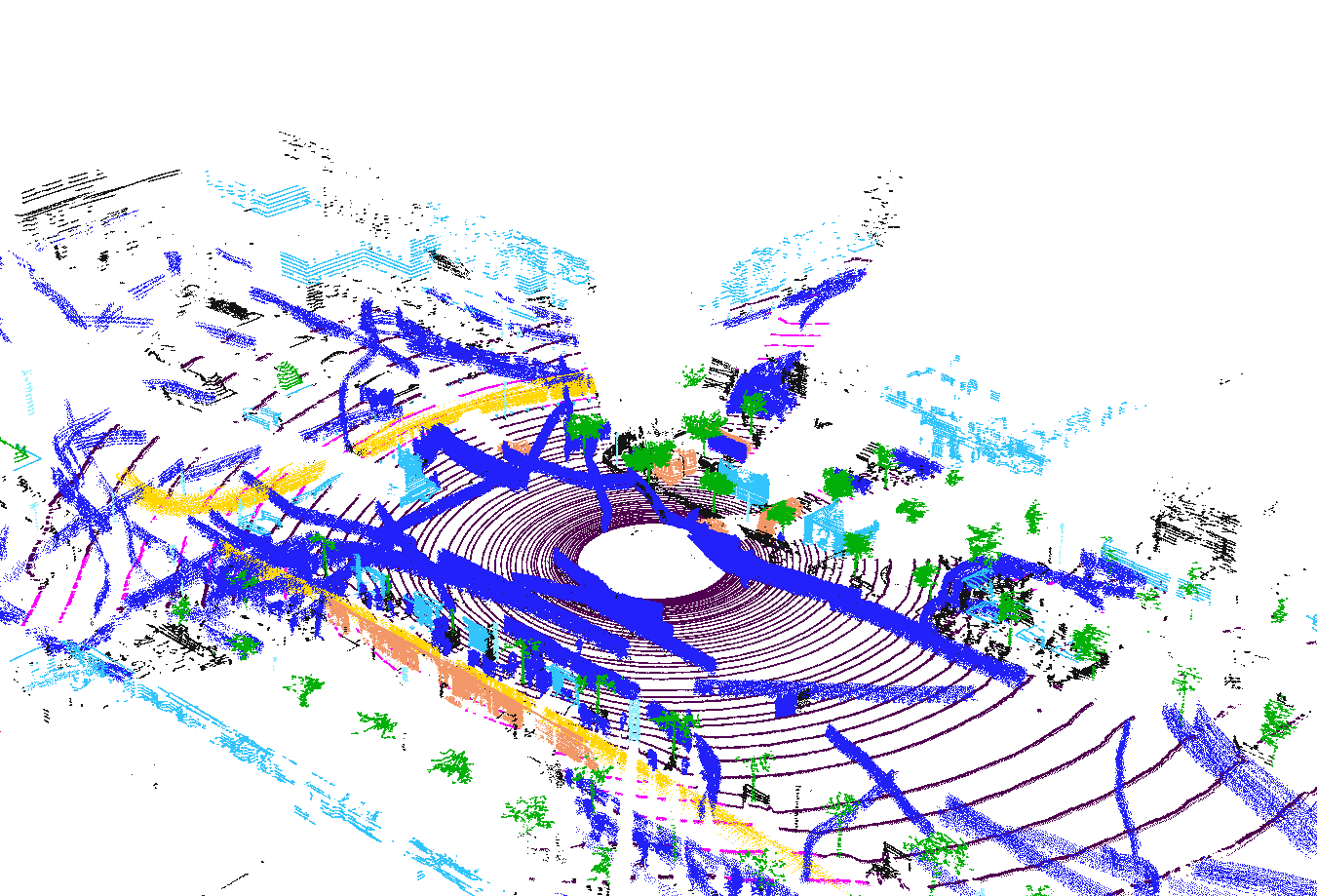} &
    \includegraphicszoomhelixnet{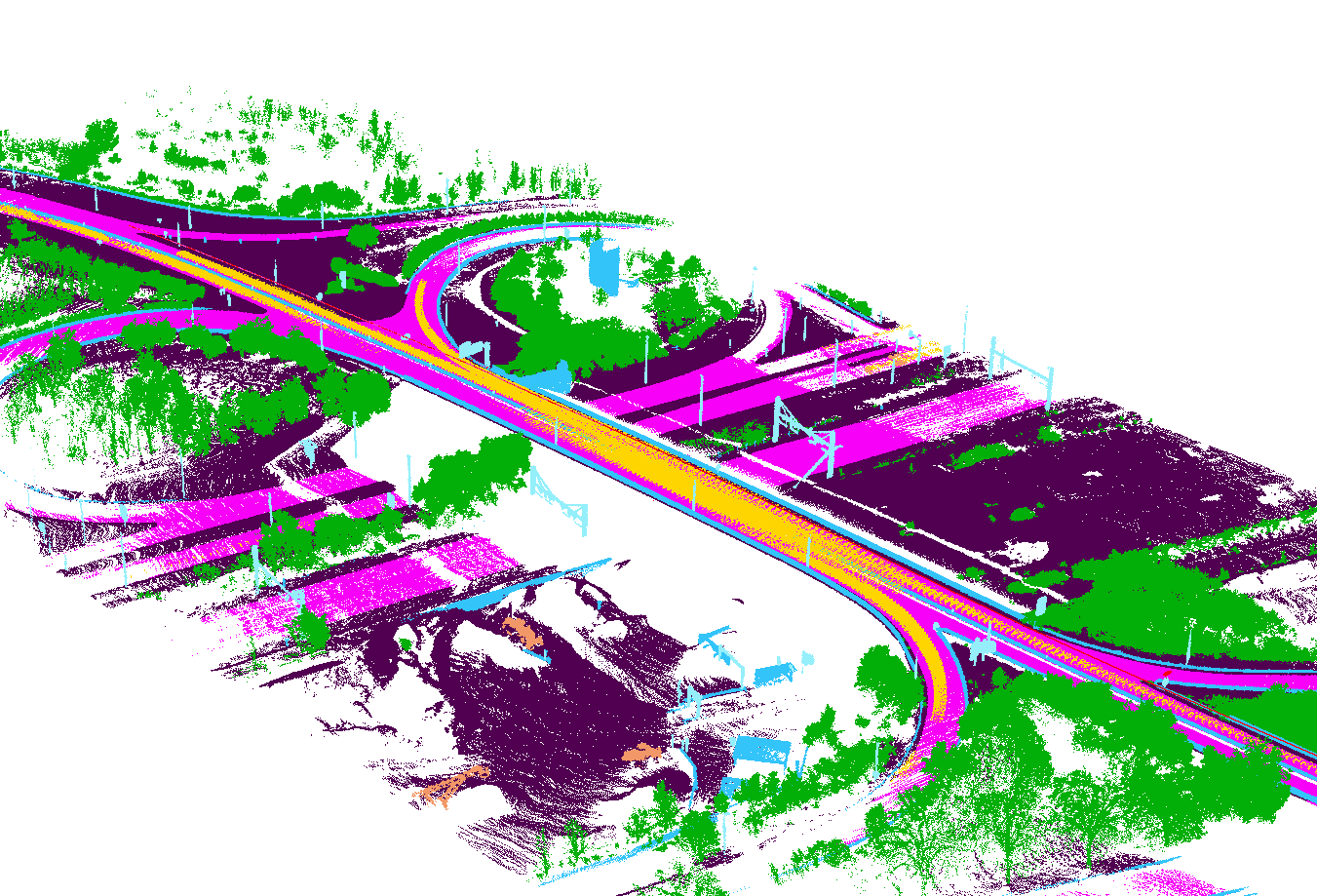} &  
    \includegraphicszoomhelixnet{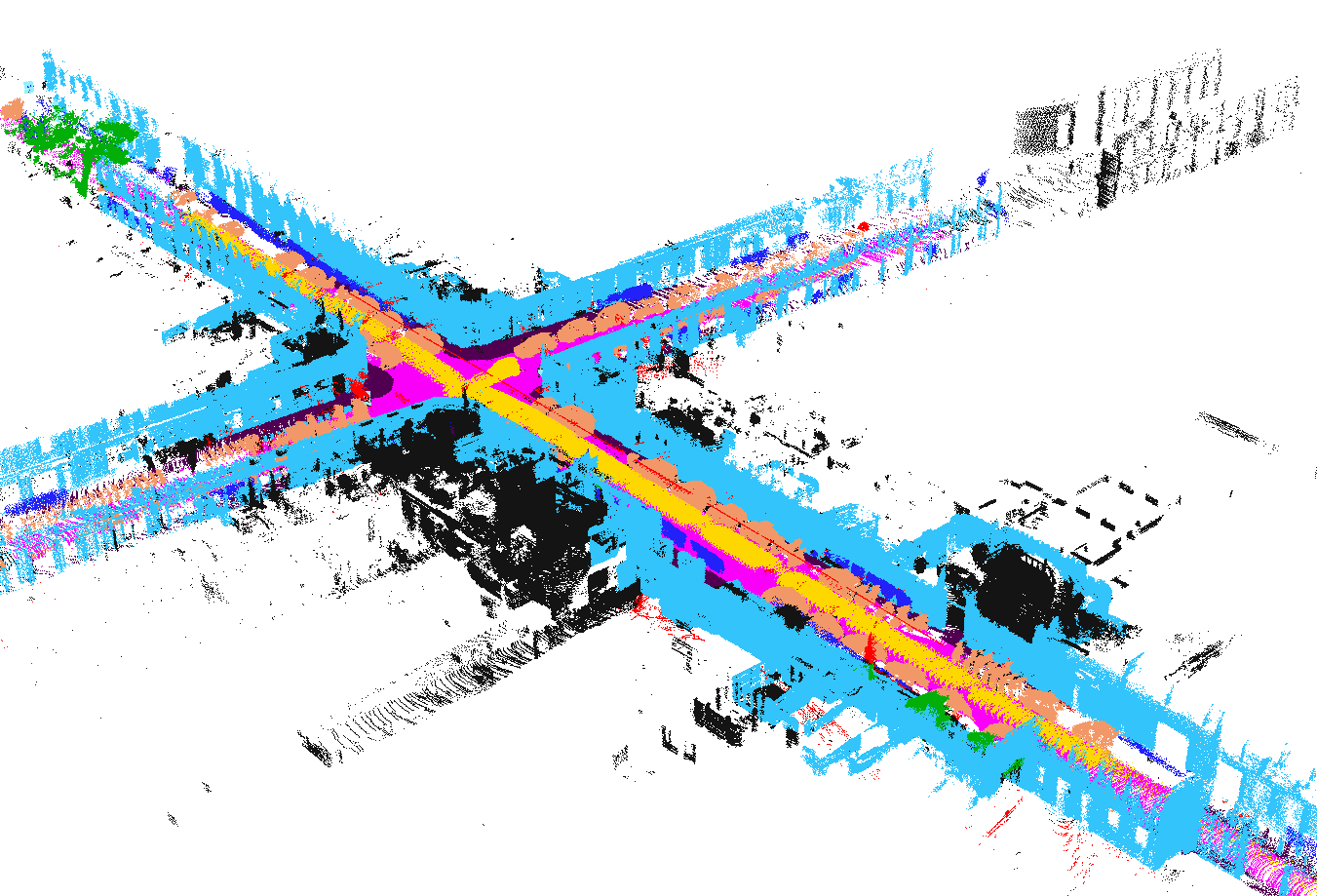} 
    \\
    &
    Seq. 1- Clermont-Ferrand &
    Seq. 5- CASQY &
    Seq. 13- Paris (5°)
\end{tabular*}

\begin{tabular*}{\textwidth}{@{}@{\extracolsep{\fill}}*{5}{l}@{}}
\classbox{class0}~{Unlabeled} & \classbox{class1}~{Road} & \classbox{class2}~{Other surface} & \classbox{class3}~{Building} & \classbox{class4}~{Vegetation} \\
\classbox{class5}~{Traffic signs} & \classbox{class6}~{Static vehicle} & \classbox{class7}~{Moving vehicle} & \classbox{class8}~{Pedestrian} & \classbox{class9}~{Artifact}
\end{tabular*}

%% file: text/FINAL/40_method.tex
\section{Helix4D: Fast LiDAR Segmentation with  Transformers}\label{sec:method}

\begin{figure*}[t]
    \centering
    \includegraphics[width=.95\textwidth, clip, trim=0 3pt 5pt 0]{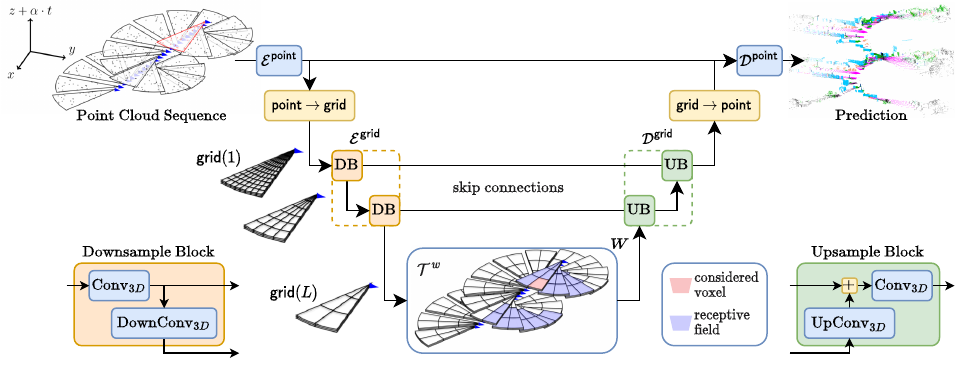}
    \caption{\textbf{Helix4D Architecture.}~
    A point sequence is split into angular slices, whose points are encoded by $\Epoi$ and pooled along a fine-grained cylindrical partition. A convolutional encoder $\Egri$ yields feature maps at lower resolutions. We apply $W$ consecutive spatio-temporal transformer blocks
    $\cT^w$ on the coarse voxels, with attention spanning across current and past slices.
    The resulting {features}
    are up-sampled to full resolution with a convolutional decoder $\Dgri$ using the encoder's maps at intermediate resolutions through skip connections. Finally, the grid features are allocated to the points, which are classified by $\Dpoi$.}
    \label{fig:pipeline}
\end{figure*}

We consider a sequence of 3D points acquired by a rotating LiDAR on a mobile platform, which we split into chronologically ordered slices of acquisition. As represented in \figref{fig:pipeline}, we process each slice with a U-Net architecture \cite{ronneberger2015u} with cylindrical convolutions \cite{zhu2021cylindrical}. At the lowest resolution, a spatio-temporal transformer network connects neighboring voxels in space and time, resulting in a large receptive field.
We first describe the construction of slices, then our cylindrical U-Net, and finally the transformer module.
\subsection{Temporal Slicing}\label{sec:slices}
Instead of processing the data frame-by-frame, we propose to split the sequence into slices covering a fixed portion of the sensor rotation, resulting in a shorter acquisition time and a lower latency.
Each point $i$ of the sequence is characterized by the angular position $\theta_i$ \emph{of the sensor head} at its exact time of acquisition. The points are sorted in chronological acquisition order \ie $\theta_i \leq \theta_j$ if $i<j$.
We partition the sequence into groups of contiguous points called slices, acquired during a portion $\Delta \theta \in ]0,2\pi]$ of a full rotation of the sensor itself. Choosing $\Delta \theta = 2\pi$ corresponds to the classic frame-by-frame setting and implies an acquisition latency of $104$ms in HelixNet or SemanticKITTI \cite{behley2019iccv}. A slice size of $\Delta \theta=2{\pi}/5$ leads to an acquisition latency of $21$ms, which is more conducive to real-time processing of driving data.
\subsection{Cylindrical U-Net}
Inspired by the Cylinder3D model~\cite{zhu2021cylindrical}, we first discretize each slice along a fine cylindrical partition $\gridi{1}$. Each point $i$ is associated with a descriptor $x^\point_i$ based on its intensity, relative position with respect to the sensor in Cartesian and cylindrical coordinates, and its offset with respect to the center of its voxels in $\gridi{1}$.  
We compute the point feature $f^\point_i$ by applying a shared Multi-Layer Perceptron (MLP) $\Epoi$ to $x^\point_i$ for all points $i$ in the slice. The resulting $f^\point_i$ are then maxpooled with respect to the voxels of $\gridi{1}$ to serve as input to a convolutional encoder $\Egri$. The network $\Egri$ is composed of sparse cylindrical convolutions~\cite{SubmanifoldSparseConvNet} and strided convolutions for downsampling. $\Egri$ produces a set of $L$ sparse feature maps $f^{\gridi{1}}, \cdots, f^{\gridi{L}}$ with decreasing resolutions:
%
\begin{alignat}{2}
    &f_i^\point &&= \Epoi\left(x^\point_i\right)\\
    f^\gridi{1}, \cdots, &f^\gridi{L} &&= \Egri\left(\maxpool\left(f^\point\right)\right)~,
\end{alignat}
where $\maxpool$ is performed with respect to $\gridi{1}$.
At the lowest resolution $\gridi{L}$, we apply the transformer-based module $\cT$ presented in the next subsection to the feature map $f^\gridi{L}$ to obtain the coarse cylindrical map $g^\gridi{L}$:
\begin{alignat}{2}
    &g^{\gridi{L}}
    &&= \cT\left(f^{\gridi{L}}\right)~.
\end{alignat}
The decoder $\Dgri$ combines cylindrical convolutions and strided transposed convolutions to map $g^\gridi{L}$ to a feature map $g^\gridi{1}$ at the highest resolution, and uses the maps $f^\gridi{L-1}, \cdots, f^\gridi{1}$ through residual skip connections.
We concatenate for each point $i$ the descriptor $g^\gridi{1}(i)$ of its voxel in $\gridi{1}$ and its point feature $f^\point_i$. Finally, the point decoder $\Dpoi$ associates a vector of class scores  $c^\point_i$ with each point $i$:
\begin{alignat}{2}
    &g^{\gridi{1}} 
    &&= \Dgri\left(g^{\gridi{L}}, f^{\gridi{L-1}}, \cdots, f^{\gridi{1}} \right)\\
    &c^\point_i 
    &&= \Dpoi\left(\left[g^{\gridi{1}}(i), f^\point_i \right]\right)~,
\end{alignat}
where $[\,\cdot\,]$ is the channelwise concatenation operator. The network is supervised by the cross-entropy and Lov\'asz-softmax \cite{berman2018lovasz} losses directly on the point prediction, without class weights.

Our approach differs from Cylinder3D \cite{zhu2021cylindrical} by relying on simple $3\times3\times3$ sparse cylindrical convolutions instead of asymmetrical convolutions and dimension-based context modeling. Furthermore, we do not use voxel-wise supervision.

Our simplified architecture results in a lighter computational and memory load, but can still learn rich spatio-temporal features thanks to the addition of the transformer module described below.
\subsection{Spatio-Temporal Transformer}
\begin{figure*}[t]
    \centering
    \input{figures/attention.tex}
    \begin{tabular*}{\textwidth}{@{}@{\extracolsep{\fill}}*{4}{l}@{}}
        \scalebox{.85}{\protect\tikz \protect\draw[green, line width=1pt] (0,0) circle (0.1) ;~spatio-temporal mask $M(v)$} &
        \scalebox{.85}{\intexttrapezium{green!50!black}~voxel $v$} &
        \scalebox{.85}{\intextsensor~sensor position} &
        \scalebox{.85}{\intextcolorscale{coolwarm}~cross-voxel attention}
    \end{tabular*}
    \caption{\textbf{Spatio-Temporal Attention.}~{We represent the spatio-temporal mask and attention score of one head of the transformer for two different voxels. The network gathers information from different frame offsets $P$ as the {sensor moves.}
    }
    }
    \label{fig:attention}
\end{figure*}
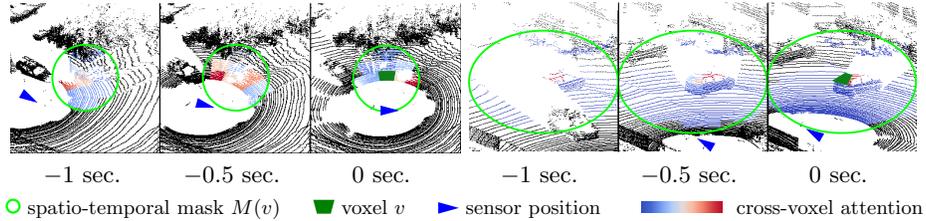

We denote by $\cV$ the set of non-empty voxels at the lowest resolution $\gridi{L}$ \emph{for all slices of the considered sequence}. 
We associate with each voxel $v$ of $\cV$ a feature $f^\voxel_v$ defined as the value of $f^{\gridi{L}}$ at $v$.
We remark that $f^\voxel$ can be ordered as a non-strictly ordered time sequence, and propose to successively apply $W$ independent transformer blocks $\cT^1, \cdots,\cT^W$ whose architecture is described below. We denote by $g^\voxel$ the resulting spatio-temporal voxel representation:
\begin{align}
g^\voxel = \cT^{{W}} \circ \cdots \circ \cT^{{1}}(f^\voxel) ~.
\end{align}
We associate each voxel $v$ of $\cV$ with the absolute position $(X_v, Y_v, Z_v)$ of its center, the release time $T_v$ of its first point, and the index $I_v$ of the sensor rotation of its corresponding slice.
In order to use a sparse attention scheme, we define for each voxel $v$ a spatio-temporal mask $M(v)$ characterized by a radius $R$ and a set of rotation offsets $P \subset \mathbb{N}$:
\begin{equation}
    M(v) = \left\{ u~\middle |~\lVert (X_v,Y_v,Z_v)-(X_u,Y_u,Z_u) \rVert <R \,,\, I_v - I_u \in P\right\}~.
\end{equation}
In the context of autonomous driving, we choose $R=6$m and $P=\{0, 5, 10\}$. With a standard rotation speed of $10$Hz, this corresponds to considering slices $0.5$ and $1$ seconds in the past along with the current one. See \figref{fig:attention} for an illustration of the receptive field and attention maps.

\paragraph{Simplified Transformer Block.} We now define a single transformer block $\cT^w$ with $H$ heads operating on a sequence of voxel features $f^\voxel$ of dimension $D$. For each head $h$ and each voxel $v$, we apply the following operations: 
\begin{itemize}[leftmargin=8mm]
\item [(i)] A single linear layer $\cL^h$ generates both a key $\key_v^h$ of dimension $K$ and a value $\Tval_v^h$ of dimension $D/H$.
\item [(ii)] For all voxels $u$ in the mask $M(v)$, we define the compatibility score $y_{u,v}^h$ as the cross-product between keys and with a learned relative positional encoding $\posenc^h(u,v)$.
\item [(iii)] The cross-voxel attention $a^h_{u,v}$ is obtained with a scaled softmax.
\item [(iv)] The values $\Tval_u^h$ of voxels in $M(v)$ are averaged into a vector $\tilde{f}_v^h$ using their respective cross-voxel attention as weights.
\item [(v)] The vectors $\tilde{f}_v^h$ are concatenated channelwise across heads and added to the input of the block to define its output.
\end{itemize}
These operations can be summarized as follows:
\begin{alignat}{3}
    \hfill
    & \key_v^h, \Tval_v^h
    & =\; & \cL^h\left(f^\voxel_v\right) \label{eq:trans}\\
    \hfill
    & y_{u,v}^h
    & =\; & \left(\key_v^h\right)^\intercal \left(\key_u^h+ \posenc^h(u,v)\right) \quad \text{for} \; u \in M(v) \label{eq:compat}\\
    \hfill
    & \left\{a_{u,v}^h\right\}_{u \in M(v)}
    & =\; & \softmax\left(\left\{y_{u,v}^h\right\}_{u \in M(v)} / \sqrt{K}\right) \label{eq:attention }\\
    \hfill
    & \tilde{f}_v^h
    & =\; & \textstyle{\sum_{u \in M(v)} a_{u,v}^h\Tval_u^{h}} \\
    \hfill
    & \cT^w(f^\voxel)_v
    & =\; & f^\voxel_v + [\tilde{f}_v^{1}, \cdots, \tilde{f}_v^{H}]~.
\end{alignat}
Our design is similar to the classical transformer architecture but uses keys as queries to save memory and computation. We also do not use feed-forward networks after averaging the values: the only learnable part of a block $\cT^w$ is its linear layers $\cL^h$ and its relative positional encoding $\posenc^h$. 

Since $g^\voxel$ only requires information about the voxels of the current and past slices, it can be computed sequentially for all slices in the order in which the sensor releases them. 
For a given slice, the voxel map $g^\gridi{L}$ for non-empty voxels is given by the values of $g^\voxel$, and set to zero otherwise.
To save computation at inference time, we store in memory the keys, values, and absolute positions of the voxels in  past slices with a fixed buffer of $\max(P)$ rotations. This allows us to allocate a large spatio-temporal receptive field to each voxel without supplementary computations. 

\paragraph{Relative Positional Encoding.} We propose to learn relative positional vectors $\posenc^{h}(u, v)$ that encode the spatio-temporal offset $(X_u, Y_u, Z_u, T_u)-(X_v, Y_v, Z_v, T_v)$ between voxels $u$ and $v$ for each transformer block $w$ independently. Inspired by the work of Wu~\etal~\cite{wu2021rethinking}, we first discretize the offsets along each dimension $\dimd \in \{X,Y,Z,T\}$ with $B_\dimd$ irregular bins. For each dimension $\dimd$ and head $h$, we learn $B_\dimd$ weight vectors of size $K$. We define the functions $\posenc^{h}_\dimd:\mathbb{R}\mapsto\mathbb{R}^K$ that map the $\dimd$-dimension of an offset to the vector associated with its corresponding bin. The positional encoding between two voxels $u$ and $v$ is the sum of the vectors corresponding to their discretized offsets in each dimension:
\begin{alignat}{4}\nonumber
        \posenc^{h}(u, v)
        =&\;\posenc^{h}_X(X_u-X_v)
        &\;+\;&\posenc^{h}_Y(Y_u-Y_v)\\
        +&\;\posenc^{h}_Z(Z_u-Z_v)
        &\;+\;&\posenc^{h}_T(T_u-T_v)~.
\end{alignat}
Relative positional encoding vectors are used directly in the calculation of the compatibility score, as given in \eqref{eq:compat}.
Additional details on positional encoding are given in the supplementary material.

%% file: figures/attention.tex
\newcommand{\PreserveBackslash}[1]{\let\temp=\\#1\let\\=\temp}
\newcolumntype{C}[1]{>{\PreserveBackslash\centering}p{#1}}
{\centering
\renewcommand{\arrayrulewidth}{0pt}
\begin{tabular}{C{.155\textwidth}C{.155\textwidth}C{.155\textwidth}C{.155\textwidth}C{.155\textwidth}C{.155\textwidth}}
    \multicolumn{3}{c}{
        \includegraphics[width=.49\textwidth, clip, trim=1.5pt 3pt 2pt 1.5pt]{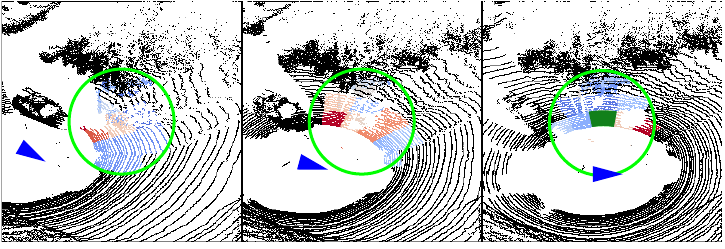}
    } & 
    \multicolumn{3}{c}{
        \includegraphics[width=.49\textwidth, clip, trim=0 3pt 0 1pt]{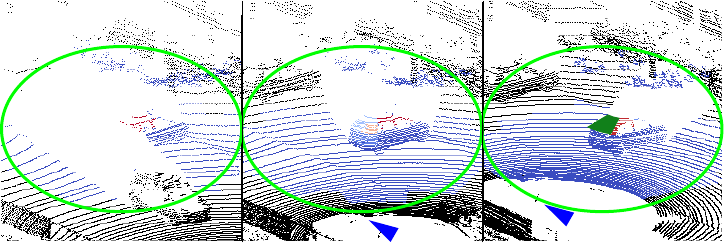}
    } \\
    $-1$~sec. & $-0.5$~sec. & $0$~sec. & $-1$~sec. & $-0.5$~sec. & $0$~sec. \\
\end{tabular}}

%% file: text/FINAL/50_applications.tex
\section{Evaluating Online Semantic Segmentation}\label{sec:experiments}
We evaluate the performance and inference time of our approach and other state-of-the-art methods in both online and frame-by-frame settings. We use our proposed dataset HelixNet and the standard SemanticKITTI dataset.

\paragraph{Online Evaluation Setting}
We aim at evaluating the real-time readiness of rotating LiDAR semantic segmentation algorithms in the context of autonomous driving. The total latency of a model is determined by its inference speed and also the time it takes to acquire its input. Operating on full frames requires at least $104$ms of acquisition, which is incompatible with realistic autonomous driving scenarios. Instead, we propose an online evaluation setting using the slices defined in \Secref{sec:slices}. By default, we use a slice size of a fifth turn of the sensor head: $\Delta \theta = 2\pi/5$, corresponding to $21$ms of acquisition.

Slices are processed sequentially. We define the inference latency of a segmentation method as the average time between the release of the last point of a slice and its segmentation. To meet the real-time requirement, inference must be faster than the acquisition of a slice. Slower processing would cause the classification to continuously fall behind.
Although thinner slices directly reduce acquisition latency, they also make the real-time requirement more strict: as a full turn must be processed in less than $104$ms, a fifth turn must be in at most $21$ms.

\paragraph{Adapting SemanticKITTI.}
SemanticKITTI~\cite{behley2019iccv,Geiger2013IJRR} contains  $43\,552$ frames along $22$ sequences of LiDAR scans densely annotated with $19$ classes. In contrast to HelixNet, SemanticKITTI is not formatted with respect to the sensor rotation and only gives the acquisition time and sensor position once per frame. To measure the latency, we make the following approximation:
(i) the fibers are assumed to be vertically aligned, meaning that the angle of the points is the same as the sensor's;
(ii) we interpolate the acquisition time of points between frames from their angular positions;
(iii) we use the acquisition time as release time.
To obtain the absolute positions of the voxels, we assume that the sensor jumps between the positions given by the camera poses for each frame. In our open-source implementation, we provide an adapted dataloader allowing methods already running on SemanticKITTI to be evaluated in the online setting with minimal adaptation.

\paragraph{Adapting Competing Methods.}
To evaluate the semantic segmentation performance and latency of other segmentation algorithms in the online setting, we process the point clouds corresponding to each slice independently and sequentially. This approach restricts the spatial receptive field to the extent of the slices. However, as the sensor moves, it is not straightforward to add past slices whose relative positions may no longer be valid. By explicitly modeling the spatio-temporal offset between voxels, Helix4D does not suffer from this limitation.
\begin{table}[t]
    \centering
    \small
    \caption{\textbf{Semantic Segmentation Results.}~
     Performance of Helix4D and competing approaches on HelixNet and on the validation set of SemanticKITTI$^\star$, in the frame-by-frame and online setting. We report the mean Intersection-over-Union (mIoU) and the inference time in ms. Methods meeting the real-time requirement are indicated with \cmark and those who do not with \xmark. $\star$ SemanticKITTI is denoted as SK. Measuring the latency on this dataset requires making non-realistic approximations about the fiber position.
    }
    \label{tab:helixformervsothers}
    \renewcommand{\arraystretch}{1}
    \begin{tabular*}{\textwidth}{@{}lr@{\extracolsep{\fill}}*{6}{c}@{}}
    \toprule
        \multirow{2}{*}{Method} & Size & \multicolumn{2}{c}{Full frame
      \tikz\fill[fill=\slicecolor, draw=none ,scale=0.25] (0,0) -- (0,0.5) arc [start angle=90, delta angle=360, radius=0.5] -- (0,0); 
    }& 104ms & \multicolumn{2}{c}{\sfrac{1}{5} frame
      \tikz\fill[fill=\slicecolor, draw=none ,scale=0.45] (0,0) -- (0,0.5) arc [start angle=90, delta angle=-72, radius=0.5] -- (0,0);
    } & 21ms\\\cline{3-5}\cline{6-8}
    
        & \textit{$\times 10^6$}                  & HelixNet  & SK$^\star$ & Inf. (ms)   & HelixNet      & SK$^\star$ & Inf. (ms) \\\midrule
        SalsaNeXt~\cite{cortinhal2020salsanext}   & 6.7     &  69.4     & 55.8     & \bf 23 \cmark & 68.2      & 55.6     & \bf 10 \cmark\\
        PolarNet~\cite{Zhang_2020_CVPR}           & 13.6    &  73.6     & 58.2     & 49 \cmark     & 72.2      & 56.9     & 36 \xmark\\
        Pan. PolarNet~\cite{Zhou2021PanopticPolarNet}           & 13.7    &  ---     & 64.5     & 50 \cmark     & ---      & 60.3     & 44 \xmark\\
        SPVNAS~\cite{tang2020searching}           & 10.8    &  73.4     & 64.7     & 73 \cmark     & 69.9      & 57.8     & 44 \xmark\\
        Cylinder3D~\cite{zhu2021cylindrical}      & 55.9    &  76.6     & \bf 66.9 & \!108 \xmark    & 75.0      & 65.3     & 54 \xmark\\
        \bf Helix4D (Ours)                    & \textbf{\oursnparams} & \bf 79.4 & 66.7     & 45 \cmark     & \bf 78.7 & \bf 66.8 & 19 \cmark\\
    \bottomrule
    \end{tabular*}
\end{table}
\begin{table}[t]
     \centering
     \small
     \caption{\textbf{HelixNet Semantic Segmentation Scores.}  We report the IoU for each class of HelixNet evaluated in the online setting with slices of $72^\circ$.}
     \label{tab:perf_classes}
     \newcommand{\hnetclassesintable}[1]{\rotatebox{35}{#1}}
    \newcommand{\PreserveBackslash}[1]{\let\temp=\\#1\let\\=\temp}
    \newcolumntype{C}[1]{>{\PreserveBackslash\centering}p{#1}}
    \newcolumntype{R}[1]{>{\PreserveBackslash\raggedleft}p{#1}}
    \newcolumntype{L}[1]{>{\PreserveBackslash\raggedright}p{#1}}
    \renewcommand{\arraystretch}{1}
     \begin{tabular}{@{}L{2.7cm}C{.85cm}C{.85cm}C{.85cm}C{.85cm}C{.85cm}C{.85cm}C{.85cm}C{.85cm}C{.85cm}C{.85cm}@{}}
        \toprule
          Method & \hnetclassesintable{Road} & \hnetclassesintable{Other surface} & \hnetclassesintable{Building} & \hnetclassesintable{Vegetation} & \hnetclassesintable{Traffic signs} & \hnetclassesintable{Static vehicle} & \hnetclassesintable{Moving vehicle} & \hnetclassesintable{Pedestrian} & \hnetclassesintable{Artifact} & Avg\\\midrule
        SalsaNeXt~\cite{cortinhal2020salsanext} &  84.4 & 76.1 & 88.7 & 70.7 & 61.4 & 58.6 & 35.1 & 68.5 & 69.7 & 68.2\\
        PolarNet~\cite{Zhang_2020_CVPR} &  86.2 & 77.9 & 91.2 & 77.9 & 63.2 & 64.8 & 35.4 & 68.1 & 84.8 & 72.2\\
        SPVNAS~\cite{tang2020searching}  & 80.5 & 77.1 & 93.0 & 81.8 & 68.0 & 60.9 & 36.9 & 71.7 & 59.0 & 69.9\\
        Cylinder3D~\cite{zhu2021cylindrical}  & 85.3 & 78.4 & 93.5 &  83.9 & 66.2 & 63.3 & 35.7 & 77.7 &  90.9 & 75.0\\
        \bf Helix4D (Ours) & \bf 87.8 & \bf 82.5 & \bf 94.0 & \bf 84.4 & \bf 68.9 & \bf 72.3 & \bf 46.4 &  \bf 78.8 & \bf 93.3 & \bf 78.7\\
        \bottomrule
     \end{tabular}
 \end{table}

We selected five segmentation algorithms with open-source implementations and trained models for SemanticKITTI. SalsaNeXt~\cite{cortinhal2020salsanext} uses range images, PolarNet~\cite{Zhang_2020_CVPR} and panoptic PolarNet~\cite{Zhou2021PanopticPolarNet} a bird's eye view polar grid, SPVNAS~\cite{tang2020searching} a regular grid, and Cylinder3D~\cite{zhu2021cylindrical} a cylindrical grid. We do not consider methods that stack frames as their structure and resulting latency is incompatible with the online setting. When using SemanticKITTI, we evaluate the provided pretrained models on the validation set. On HelixNet, we retrain the models from scratch using the procedure of their official repository. We removed all test-time augmentations that resulted in prohibitive inference time. All methods are evaluated on the same workstation using a NVIDIA TESLA V100 32Go GPU.

\begin{figure}[t]
    \centering
    \input{tikz/time_and_miou_semkitti.tex}
	\caption{{\bf Influence of Slice Size.}~{We plot the processing time (left, in ms) and precision (right, in mIoU) of different methods with respect to the considered size of slices, estimated on the validation set of SemanticKITTI~\cite{behley2019iccv}. 
	Methods whose inference time is slower than the acquisition time of the slice (red shaded area) do not meet the real time requirement.
	}}
    \label{fig:main}
\end{figure}
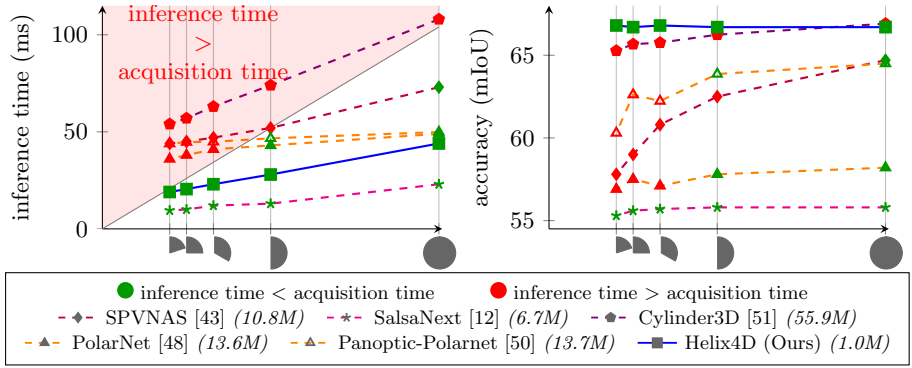

\paragraph{Analysis.}
In \tabref{tab:helixformervsothers}, we report performance in frame-by-frame and online setting with slices of $72^\circ$, for Helix4D and competing methods, for HelixNet and SemanticKITTI. 
We observe that Helix4D yields state-of-the-art accuracy, with mIoU scores only matched by Cylinder3D~\cite{zhu2021cylindrical}. However, Cylinder3D is $50$ times larger in terms of parameters and twice slower, not meeting the real-time requirement even in the full frame setting. As reported in \tabref{tab:perf_classes}, 
distinguishing moving vehicles in HelixNet is particularly difficult. Our approach even largely outperforms Panoptic PolarNet despite this method using instance annotation as supervision, preventing us from evaluating on HelixNet. Helix4D yields significantly improved scores thanks to its larger spatio-temporal receptive fields: $14$m and $1000$ms \vs $8$m and $21$ms for Cylinder3D for a fifth rotation.

In the online setting, only two approaches meet the real-time requirement: SalsaNeXt~\cite{cortinhal2020salsanext} and Helix4D. Our approach outperforms SalsaNeXt by over $10$ mIoU points in both the full frame and the on-line settings. In short, Helix4D is as accurate as the largest and slowest models with an inference speed comparable to that of the fastest and less accurate models. The total latency (acquisition plus inference time) of our model evaluated online is $40$ms ($21+19$ms), and reaches the same performance as Cylinder3D evaluated on full frame with a latency of $212$ms ($104+108$ms), an acceleration of more than $5$ folds.

In \figref{fig:main}, we report the inference time and mIoU for different slice sizes. Due to various overheads, the inference time appears in an affine relationship with the size of slices, making the real-time requirement stricter for smaller slices. Due to its very design, the performance of Helix4D is not affected by the slice size. In contrast, competing methods perform worse with smaller slices.

\begin{table}[t]
    \centering
    \small
    \caption{\textbf{Ablation Study.} We report the speed and accuracy of several modification of our Helix4D on the validation set of SemanticKITTI.
    \label{tab:ablation}
    }
    \renewcommand{\arraystretch}{1}
    \begin{tabular*}{\textwidth}{@{}l@{\extracolsep{\fill}}*{5}{c}@{}}\toprule
        \multirow{2}{*}{Method} & Size & \multicolumn{1}{c}{Full Frame
      \tikz\fill[fill=\slicecolor, draw=none ,scale=0.25] (0,0) -- (0,0.5) arc [start angle=90, delta angle=360, radius=0.5] -- (0,0);}
    & 104ms & \multicolumn{1}{c}{\sfrac{1}{5} Frame
      \tikz\fill[fill=\slicecolor, draw=none ,scale=0.45] (0,0) -- (0,0.5) arc [start angle=90, delta angle=-72, radius=0.5] -- (0,0);}
    & 21ms\\\cline{3-4}\cline{5-6}
    
         & $\times 10^3$ & mIoU & Inf. (ms) & mIoU & Inf. (ms) \\\midrule
        Helix4D          & 985 & \textbf{66.7} & 45 \cmark  & {\textbf{66.8}} & {{19}} \cmark\\
        \enumref{abla:asymconv} Asymmetric Convolutions   & 1171& 66.6  & 56 \cmark & 66.6            & 31 \xmark\\
        \enumref{abla:cunet} Cylindrical U-Net      & 985 & 58.6 & \textbf{22} \cmark  & 60.2   & \bf16 \cmark\\
        \enumref{abla:sbs} Slice-by-Slice            & 985 & 62.9 & 29 \cmark  & 62.6            & 19 \cmark\\
        \enumref{abla:wqueries} w. Queries                & 993 & 65.2 & 45 \cmark  & 64.8            & 20 \cmark\\
        \enumref{abla:wope} w/o. Positional Encoding  & 983 & 64.3 & 41 \cmark  & 64.1   & 18 \cmark\\
        \enumref{abla:tiny} Helix4D Tiny     & 306 & 65.3 & 45 \cmark  & 64.9            &  17 \cmark\\
    
    \bottomrule
    \end{tabular*}
\end{table}

\paragraph{Ablation Study}
We assess on SemanticKITTI the impact of different design choices by evaluating several alterations of our method, reported in \tabref{tab:ablation}.
\begin{enumerate}[label=\bf(\alph*), topsep=0cm, wide, labelwidth=!, labelindent=0pt]
    \item {\bf Asymmetric Convolutions:}\label{abla:asymconv} we replace the $3\times3\times3$ convolutions in our U-Net with the convolution design proposed by Cylinder3D~\cite{zhu2021cylindrical}. We did not observe a significant change in performance and an increase in run-time of $50\%$, failing the real-time requirement for slices of $72^\circ$.
    \item {\bf Cylindrical U-Net:}\label{abla:cunet} we replace the transformer by a $1\times1\times1$ convolution on the voxels of the lowest resolution. We observe a slight decrease in run-time and a significant drop of over $6$ mIoU points. This result shows that the transformer is able to learn meaningful spatio-temporal features at low resolution.
    \item {\bf Slice-by-Slice:}\label{abla:sbs} we restrict the mask $M(v)$ of each voxel to its current slice. This reduction in the temporal receptive field results in a drop of $4$ mIoU points, without any appreciable acceleration.
    \item {\bf w. Queries:}\label{abla:wqueries} we modify our simplified transformer to associate a query for each voxel along with keys and values, and use key-queries compatibilities. This does not affect the run-time and slightly decreases the performance.
    \item {\bf w/o. Positional Encoding:}\label{abla:wope} {we remove the relative positional encoding $\posenc$ in the calculation of compatibilities in equation \eqref{eq:compat}.} This leads to a slightly decreased run time, but decreases performance by more than $2.5$ points. This illustrates the advantage of explicitly modeling the spatio-temporal voxel offsets.
    \item {\bf Helix4D Tiny:}\label{abla:tiny} we replace the learned pooling in our U-Net with maxpools and use narrower feature maps for a total of $306k$ parameters. This method only performs two points under Helix4D with a third of its parameters.
\end{enumerate}

%% file: tikz/time_and_miou_semkitti.tex
\centering

\pgfplotsset{
  /pgfplots/xlabel near ticks/.style={
     /pgfplots/every axis x label/.style={
        at={(ticklabel cs:0.5)},anchor=near ticklabel
     }
  },
  /pgfplots/ylabel near ticks/.style={
     /pgfplots/every axis y label/.style={
        at={(ticklabel cs:0.5)},rotate=90,anchor=near ticklabel}
     }
  }

\newcommand*{\radius}{.035\linewidth}
\newcommand*{\axissize}{.735\linewidth}
\newcommand*{\subfigurewidth}{.5\textwidth}

\begin{subfigure}{\subfigurewidth}

\newcommand*{\FrameAcqTime}{104}
\newcommand*{\MaxTime}{115}
\newcommand*{\HeightProportion}{.8}

\newcommand*{\labelxshift}{0pt}
\newcommand*{\labelyshift}{0pt}
\newcommand*{\miouscale}{0}

\begin{tikzpicture}

    \begin{axis}[
		ylabel={\small inference time (ms)},
		ylabel near ticks,
        xlabel near ticks,
		axis y line=left,
		axis x line=bottom,
		xtick = {20,25,33,50,100}, xmin = 0, xmax = 101,
		xticklabel=\empty,
		ytick ={0,50,100,150,200,250}, ymin = 0, ymax = \MaxTime,
		xmajorgrids,
		legend pos=south west,
		width=\linewidth,
		height=14em,
		at={(0, 1.5*\radius)}
	]
	
	\addinferenceoverslice{Cylinder3D~\cite{zhu2021cylindrical}}{results/cylinder3d.txt}{\colorcthreed}{\patternthreed}{\markthreed}
	\addinferenceoverslice{PolarNet~\cite{Zhang_2020_CVPR}}{results/polarnet.txt}{\colorpolarnet}{\patternpolarnet}{\markpolarnet}
	\addinferenceoverslice{SPVNAS~\cite{tang2020searching}}{results/spvnas.txt}{\colorspvnas}{\patternspvnas}{\markspvnas}
	\addinferenceoverslice{SalsaNext~\cite{cortinhal2020salsanext}}{results/salsanext.txt}{\colorsalsanext}{\patternsalsanext}{\marksalsanext}
	\addinferenceoverslice{Panoptic-Polarnet~\cite{Zhou2021PanopticPolarNet}}{results/panopolar.txt}{\colorpanopolar}{\patternpanopolar}{\markpanopolar}
	\addinferenceoverslice{Ours}{results/ours_temp.txt}{\colorours}{\patternours}{\markours}

	\addplot[color=black!50,mark=,
	    ] coordinates {
		(0, 0)
		(100, \FrameAcqTime)
	};
            
    \addplot [draw=none,mark=,color=red,fill=red, 
	    fill opacity=0.1]coordinates {
            (0, 0)
            (100, \FrameAcqTime)
            (100, \MaxTime)
            (0, \MaxTime)};
            
    \node[color=red] at (axis cs: 30, 95) {\begin{tabular}{c}inference time\\$>$\\acquisition time\end{tabular} };
		
	\legend{};
	\end{axis}
	\displayslices

\end{tikzpicture}
\end{subfigure}\hfill
\begin{subfigure}{\subfigurewidth}

\begin{tikzpicture}
    \displayslices
    
    \begin{axis}[
		ylabel={\small accuracy (mIoU)},
		ylabel near ticks,
        xlabel near ticks,
		axis y line=left,
		axis x line=bottom,
		xtick = {20,25,33,50,100}, xmin = 0, xmax = 101,
		xticklabel=\empty,
		ytick ={50, 55, 60, 65}, ymax = 68, ymin = 54.5,
		xmajorgrids,
		legend pos=south west,
		width=\linewidth,
		height=14em,
		at={(0, 1.5*\radius)}
	]
	
	\addmiouoverslice{Cylinder3D~\cite{zhu2021cylindrical}}{results/cylinder3d.txt}{\colorcthreed}{\patternthreed}{\markthreed}
	\addmiouoverslice{PolarNet~\cite{Zhang_2020_CVPR}}{results/polarnet.txt}{\colorpolarnet}{\patternpolarnet}{\markpolarnet}
	\addmiouoverslice{SPVNAS~\cite{tang2020searching}}{results/spvnas.txt}{\colorspvnas}{\patternspvnas}{\markspvnas}
	\addmiouoverslice{SalsaNext~\cite{cortinhal2020salsanext}}{results/salsanext.txt}{\colorsalsanext}{\patternsalsanext}{\marksalsanext}
	\addmiouoverslice{Panoptic-Polarnet~\cite{Zhou2021PanopticPolarNet}}{results/panopolar.txt}{\colorpanopolar}{\patternpanopolar}{\markpanopolar}
	
	\addmiouoverslice{Ours}{results/ours_temp.txt}{\colorours}{\patternours}{\markours}

	\addplot [draw=none,mark=,color=red,fill=red, 
	    fill opacity=0.05]coordinates {
            (0, 0)
            (0, 70)
            (-200, 70)
            (-200, 0)};
            
	\legend{};
	\end{axis}
	
\end{tikzpicture}
\end{subfigure}

\begin{tikzpicture}
	\node[draw]{
	\scalebox{.87}{
	\begin{tabular}[width=.9\textwidth]{c}
	    \begin{tabular*}{\textwidth}{@{}@{\extracolsep{\fill}}*{4}{l}@{}}
            &\scalebox{.95}{\protect\tikz \protect\draw[\okgreen, line width=3pt, fill=\okgreen] (0,0) circle (0.08) ;~inference time $<$ acquisition time} &
            \scalebox{.95}{\protect\tikz \protect\draw[red, line width=3pt, fill=red] (0,0) circle (0.08) ;~inference time $>$ acquisition time}
        \end{tabular*}
        \\
        \addteaserlegend{SPVNAS~\cite{tang2020searching}}{10.8}{\markspvnas}
	    \addteaserlegend{SalsaNext~\cite{cortinhal2020salsanext}}{6.7}{\marksalsanext}
	    \addteaserlegend{Cylinder3D~\cite{zhu2021cylindrical}}{55.9}{\markthreed}
        \\
	    \addteaserlegend{PolarNet~\cite{Zhang_2020_CVPR}}{13.6}{\markpolarnet}
	    \addteaserlegend{Panoptic-Polarnet~\cite{Zhou2021PanopticPolarNet}}{13.7}{\markpanopolar}
        \addteaserlegend{Helix4D (Ours)}{\oursnparams}{\markours}
    \end{tabular}}
	};
\end{tikzpicture}

%% file: text/FINAL/90_conclusion.tex
\section{Conclusion}\label{sec:conclusion}
In this paper, we introduced a novel online inference setting for the semantic segmentation of sequences of rotating liDAR 3D point clouds. Our proposed large-scale dataset HelixNet contains specific sensor information that allows a rigorous evaluation of the performance and latency of segmentation methods in our online setting. We also introduced Helix4D, a transformer-based network specifically designed for online segmentation, achieving state-of-the-art results with a fraction of the latency and parameters of competing methods.
We hope that our open-source dataset and implementation will encourage the evaluation of future semantic liDAR segmentation methods in more realistic settings and help to bridge the gap between academic work on 3D perception and the operational constraints of autonomous driving.

%% file: text/91_acknowledgements.tex
\paragraph*{Acknowledgements}This work was supported in part by ANR project READY3D ANR-19-CE23-0007 and was granted access to the HPC resources of IDRIS under the allocation 2022-AD011012096R1 made by GENCI. The point cloud sequences of HelixNet were acquired during the Stereopolis II project~\cite{stereopolis}. HelixNet was annotated by \href{https://www.futurmap.com/en/our-service/}{FUTURMAP}. We thank Zenodo for hosting the dataset. We thank François Darmon, Tom Monnier, Mathis Petrovich and Damien Robert for inspiring discussions and valuable feedback.

%% file: text/FINAL/99_suppmat.tex
In this supplementary material, we provide the exact implementation of Helix4D~(\Secref{sec::implementation_details}), details about HelixNet~(\Secref{sec:details_helixnet}), and a discussion about the evaluation of online semantic segmentation~(\Secref{sec:evalonline}). Code, data and interactive visualizations are available at: {\tt{\url{https://romainloiseau.fr/helixnet}}}.

\subsection{Helix4D Implementation Details}\label{sec::implementation_details}

\begin{figure*}[h]
    \centering
    \includegraphics[width=\textwidth]{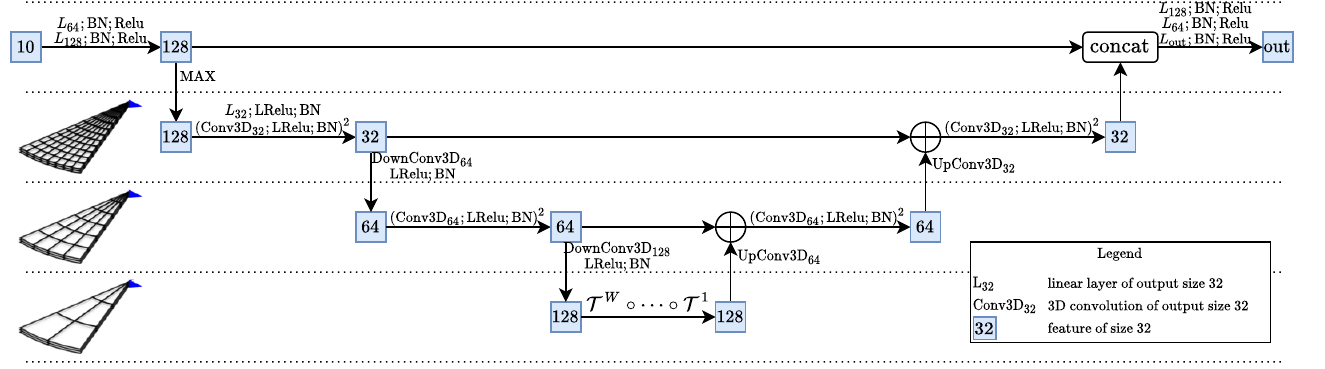}
    \caption{\textbf{Helix4D Detailled Architecture.}~Abstract version of the architecture showing all layers in $\Epoi$, $\Egri$, $\Dgri$ and $\Dpoi$ given for reproductibility.}
    \label{fig:pipeline_detailled}
\end{figure*}

\paragraph{Architecture.}
We report here the detailed architecture of the point and voxels encoders and decoders $\Epoi$, $\Egri$, $\Dgri$, $\Dpoi$ in \figref{fig:pipeline_detailled}.

We use $L=3$ levels of cylindrical grids with decreasing resolution $(\delta^l r, \delta^l \theta, \delta^l z)$. Our goal is to sufficiently decrease the resolution such that the number of nonempty voxels in $\gridi{L}$ remains manageable by the spatio-temporal transformer. We chose the following resolutions:
$\gridi{1}$ with 
$\delta^1 r = 20\text{cm}, 
\delta^1 \theta=1.33^\circ, 
\delta^1 z= 16.6\text{cm}$,
$\gridi{2}$ is of 
$\delta^1 r = 60\text{cm}, 
\delta^1 \theta=4^\circ, 
\delta^1 z= 33.3\text{cm}$, 
and $\gridi{3}$ is of 
$\delta^1 r = 180\text{cm}, 
\delta^1 \theta=8^\circ, 
\delta^1 z= 66.6\text{cm}$.
This configuration corresponds to an average of around $2500$ non empty voxels per rotations at the lowest resolution $\gridi{3}$.

$\Epoi$ and $\Dpoi$ are simple MLPs operating directly on point features, and applied in parallel to all points of a slice.
As shown in \figref{fig:pipeline_detailled}, $\Egri$ has two levels operating at different resolution in a U-Net fashion. The first level operates at the resolution of $\gridi{1}$ and is composed of two $3\times 3 \times 3$ convolutions and a strided convolution
with kernel size $[3,3,2]$ and stride $[3,3,2]$, yielding a map of resolution $\gridi{2}$.
The second level operates at the level of $\gridi{2}$ and is composed of two $3\times 3 \times 3$ convolutions and a strided convolution
with kernel size $[3,2,2]$ and stride $[3,2,2]$, yielding a map of resolution $\gridi{3}$.

The grid decoder has a symmetric architecture and uses strided transposed convolution to increase the resolution of its feature maps. The intermediate maps of $\Egri$  are simply added to the upsampled maps of $\Dgri$ at the corresponding resolution to form \emph{residual skip-connections}. 

In our spatio-temporal transformer $\cT$, we use $W=2$ successive transformer blocks, $H=4$ heads, an input size of $D=128$, and a key size of $K=8$. The masks are defined by the rotation offset $P=\{0,5,10\}$ and a radius of $R=6$m.

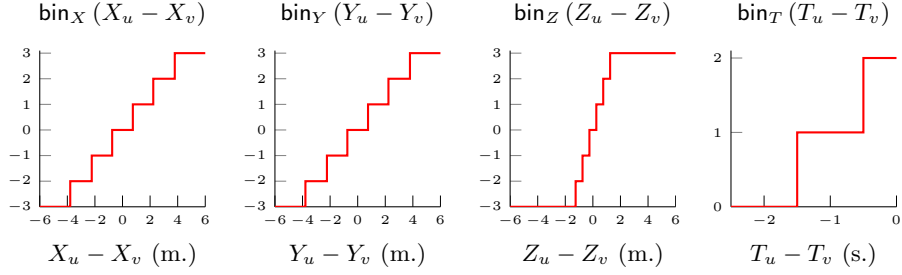
\begin{figure*}[t]
    \centering
    \input{tikz/buckets.tex}
    \caption{\textbf{Relative Positional Encoding Bins.}~{Each of the four dimension are split into irregular bins inspired by Wu~\etal~\cite{wu2021rethinking}.}}
    \label{fig:tikzbuckets}
\end{figure*}

\paragraph{Relative Positional Encoding.}
We use $B_X=B_Y=B_Z=7$ and $B_T=\text{Card}(P)=3$ irregular bins for the relative positional encoding.
For an offset $x$ along dimension $d\in\{X,Y,Z,T\}$, the corresponding bin for the pair of voxels $u, v$ is given by the following piecewise function \cite[Eq~18]{wu2021rethinking}:
\begin{align}
\mathsf{bin}_d(x)=
\begin{cases}
\lfloor x / \rho_d\rceil 
&\text{if}\; |x|\leq\alpha\rho_d \\
 \text{sign}(x)\times \min\left(\beta,\left\lfloor\alpha+\frac{\ln(|x|/(\alpha\rho_d))}{\ln(\gamma/\alpha)}(\beta-\alpha) \right\rceil\right)
 & \text{if}\; |x|>\alpha\rho_d~,
 \end{cases}
\end{align}
\noindent where $\lfloor\,\cdot\,\rceil$ denotes the rounding operation, $\alpha=2$, $\beta=3$, and $\gamma=1.25$, and $\rho_X=\rho_Y=1.5$m, $\rho_Z=0.5$m, and $\rho_T=5\times104$ms.  See \figref{fig:tikzbuckets} for a visual representation of bins.

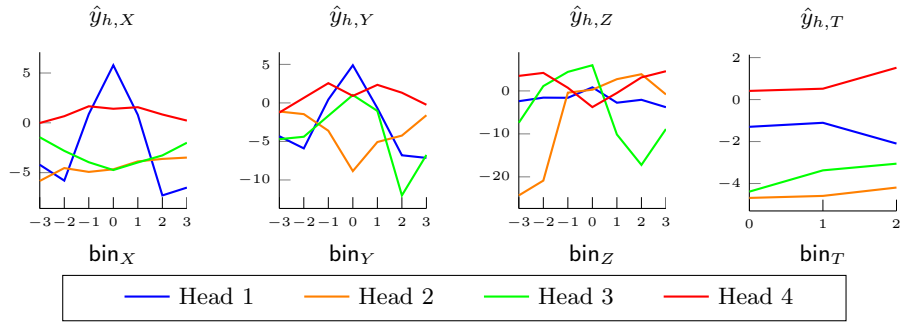
\begin{figure*}[t]
    \centering
    \input{tikz/heads.tex}
    \caption{{\bf Positional Encoding.} We plot the average compatibility with respect to the offset learned for the second encoder block of Helix4D within different dimensional bins. The compatibility is the cross product between voxel keys and the positional encoding, averaged over $3$ rotations of HelixNet (see  \eqref{eq:binhead}).}
    \label{fig:heads}
\end{figure*}

We represent in \figref{fig:heads} the compatibility $\hat{y}_{h,d}(b)$ between keys and relative positional encodings within each bin $b$, head $h$ and dimension $d\in\{X, Y, Z, T\}$ averaged over $3$ rotations: 
\begin{align}
    \hat{y}_{h,d}(b)
    &=
    \frac{1}{\text{Card}\left(\cV_d({b})\right)}\sum_{(v,u)\in\cV_d(b)} \left(\key_v^h\right)^\intercal \left(\posenc^h(u,v)\right)~, \label{eq:binhead}
\end{align}
\noindent where $\cV_d({b})=\{(v, u)\in\cV~|~\mathsf{bin}(d_u-d_v)=b\}$. We observe differentiated behaviors between the  heads: Head~1 focuses on near voxels, Head~2 and Head~3 focus on the area above/below the voxel, respectively, and Head~4 focuses on the current frame. 

\paragraph{Training.} We use the ADAM~\cite{kingma2014adam} optimizer with a weight decay of $0.01$, except for the positional encoding parameters which are learned without weight decay. We use a learning rate of $0.001$ and decrease it to $0.0001$
when the validation mIoU does not increase for consecutive 3 epochs.

\paragraph{Data Augmentation.} During training, we use augmentation to improve the generalisation capabilities of the network: flipping of the $xy$ 2D plane, rotation around the $z$ axis, anisotropic scaling generated by a uniform distribution $\mathcal{U}[0.95,1.05]$, and translation generated by a random normal $\mathcal{N}(0, 0.2m)$. Since our goal is to evaluate real-time semantic segmentation, we do not use test-time augmentation schemes, even though it could possibly improve the overall performance.

\paragraph{Real-Time Requirements.} To better represent the limitation of embarked hardware, we ran Helix4D on a NVIDIA-2080Ti and observed only a $1$ms increase in run time. In particular, Helix4D can still run in real time for slices of $72^\circ$.

\subsection{Details on HelixNet}\label{sec:details_helixnet}
\begin{figure}
    \centering
    \subcaptionbox{France\label{fig:seq:fr}}{
        \includegraphics[height=.36\textwidth]{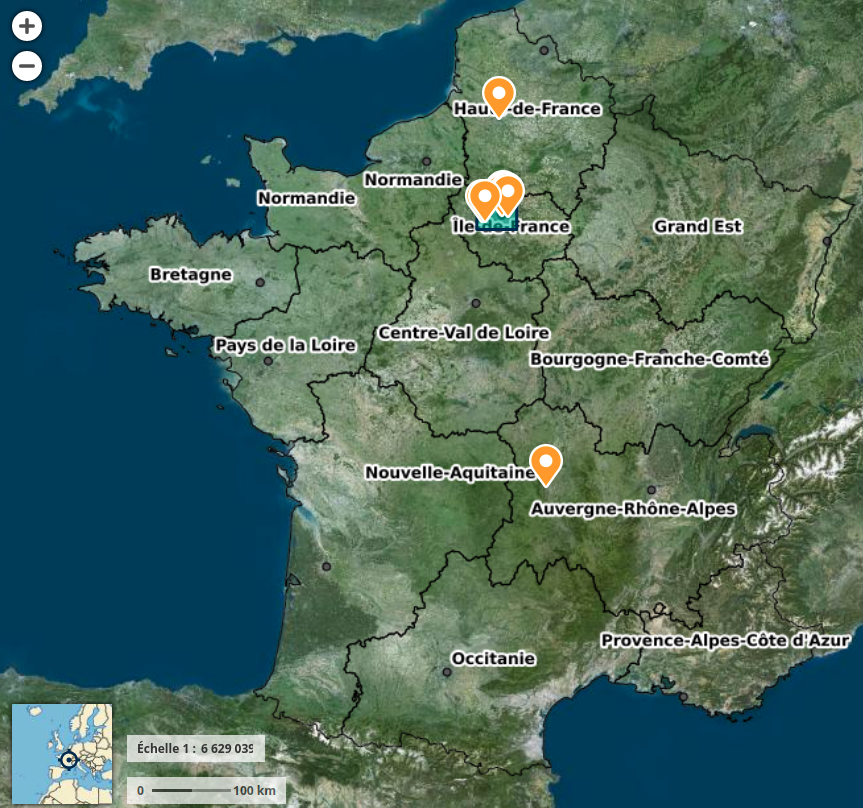}
    }
    \hfill
    \subcaptionbox{Paris agglomeration\label{fig:seq:idf}}{
        \includegraphics[height=.36\textwidth]{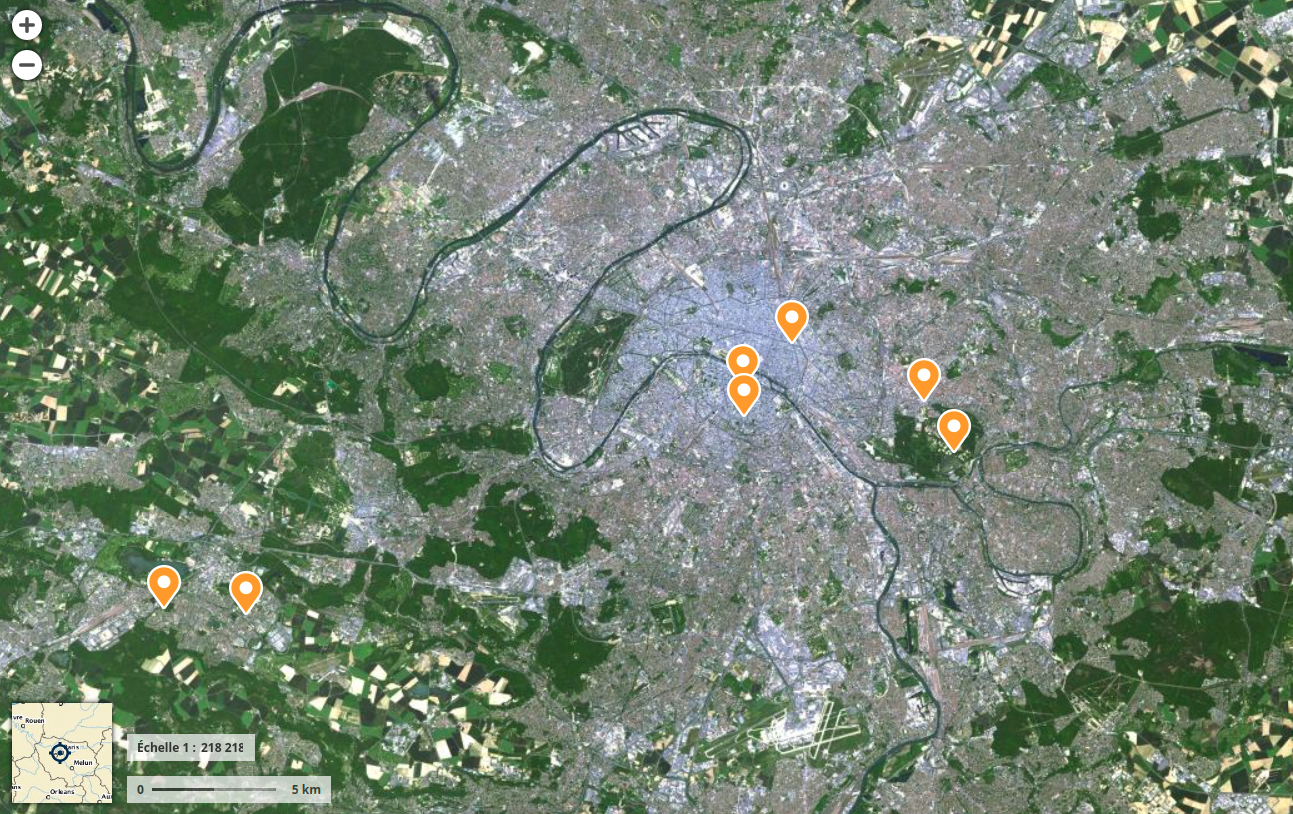}
    }
    \caption{{\bf Localization of the Sequences.} HelixNet's data were acquired in 6 different cities with 4 of them in the Paris agglomeration, spanning a large variety of landscapes and urban configuration.}
    \label{fig:france}
\end{figure}
 We report the localization of the sequences of HelixNet in detail in \figref{fig:france}. The different classes used in HelixNet are defined as followed:
 \begin{itemize}
     \item {\bf Road:} refers to the main driving lanes, not just any impervious surfaces. For example, parking spots and pedestrian alleys are excluded.
     \item {\bf Other Surface:} surface other than the road. Includes parks, grass, construction sites, parking spots, pedestrian plazas, and so on.
     \item {\bf Buildings:} comprise housings and monuments, but also urban hardscape such as bus stops, posts and poles, street fences, benches, street lights, and so on.
     \item {\bf Vegetation:} trees, shrubs, and bushes.
     \item {\bf Road Signs:} refers exclusively to hardscapes that present driving-related information: traffic signs, traffic lights, etc.
     \item {\bf Static Vehicles:} these vehicles can be parked, or simply stopped in traffic or at an intersection.
     \item {\bf Moving Vehicles:} refers to any vehicle (car, truck, bike, motorbike, construction engine) with detectable movement, assessed with a $15$s sliding time window.
     \item {\bf Pedestrians:} person travelling on foot. 
     \item {\bf Artifact:} acquisition artifacts, does not correspond to a physical impact point. This can be due to reflective surfaces or errors in the vehicle pose estimation among other reasons.
     \item {\bf Unannotated:} the human annotators could not confidently associate a label to some points due to low density or an ambiguous configuration. This is not a predicted class, and is not taken into account in the metrics.
     \item {\bf No-Echo:} some pulses never receive an echo, for example, when the impact points are too far away from the sensor. These points are affected $r=0$, and are not taken into account in the computation of metrics. Keeping these points allows us to easily convert the point cloud sequence into a sequence of range images by simply reshaping the corresponding tensor.
 \end{itemize}
\subsection{Evaluating Online Semantic Segmentation}\label{sec:evalonline}
\begin{figure}[t]
    \centering
    \begin{tabular*}{\textwidth}{@{}@{\extracolsep{\fill}}*{3}{c}@{}}
    \multicolumn{3}{c}{\begin{subfigure}[t]{.99\textwidth}
    \centering
        \includegraphics[width=\textwidth]{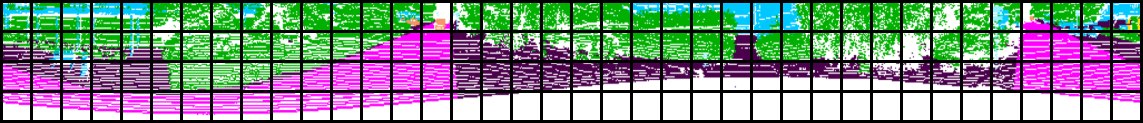}
        \subcaption{Range image~\cite{cortinhal2020salsanext}}
        \label{fig:partitions:salsa}
    \end{subfigure}}
    \\
    \begin{subfigure}[t]{.33\textwidth}
    \centering
        \includegraphics[width=\textwidth]{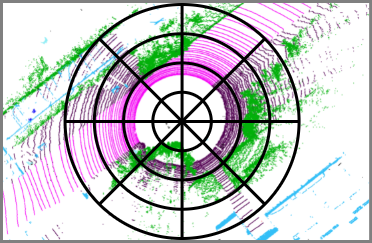}
        \subcaption{2D polar grid~\cite{Zhang_2020_CVPR} }
         \label{fig:partitions:polar}
    \end{subfigure} 
    &
    \begin{subfigure}[t]{.33\textwidth}
    \centering
        \includegraphics[width=\textwidth]{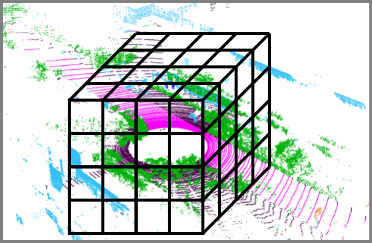}
        \subcaption{3D Euclidean grid~\cite{tang2020searching} }
         \label{fig:partitions:spvnas}
    \end{subfigure}
    &
    \begin{subfigure}[t]{.33\textwidth}
    \centering
        \includegraphics[width=\textwidth]{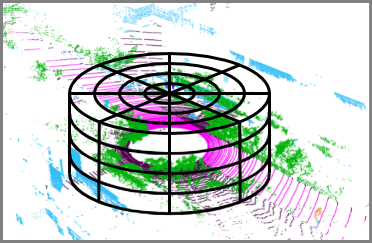}
        \subcaption{3D cylinder grid~\cite{zhu2021cylindrical} }
         \label{fig:partitions:cylinder}
    \end{subfigure}\\
    \end{tabular*}
    \caption{\textbf{Data Partitioning Schemes.}~
    We represent commonly used partitioning: \Subref{fig:partitions:salsa} SalsaNeXt~\cite{cortinhal2020salsanext} uses range images,
    \Subref{fig:partitions:polar} PolarNet~\cite{Zhang_2020_CVPR} uses a bird-eye-view polar partition,
    \Subref{fig:partitions:spvnas} 
    SPVNAS~\cite{tang2020searching}
    uses a classic regular 3D grid,
    \Subref{fig:partitions:cylinder}
    Cylinder3D~\cite{zhu2021cylindrical}
    uses a cylinder grid. Grids not to scale.}
    \label{fig:partitions}
\end{figure}

\paragraph{Competing methods.} We chose four segmentation algorithms with open-source implementations and trained models for SemanticKITTI. SalsaNeXt~\cite{cortinhal2020salsanext} uses range images, PolarNet~\cite{Zhang_2020_CVPR} a bird's-eye view polar partition, SPVNAS~\cite{tang2020searching} a regular grid, and Cylinder3D~\cite{zhu2021cylindrical} cylindrical partition. See \figref{fig:partitions} for a representation of each method's partition.

\begin{figure}
    \centering
    \includegraphics[width=\textwidth]{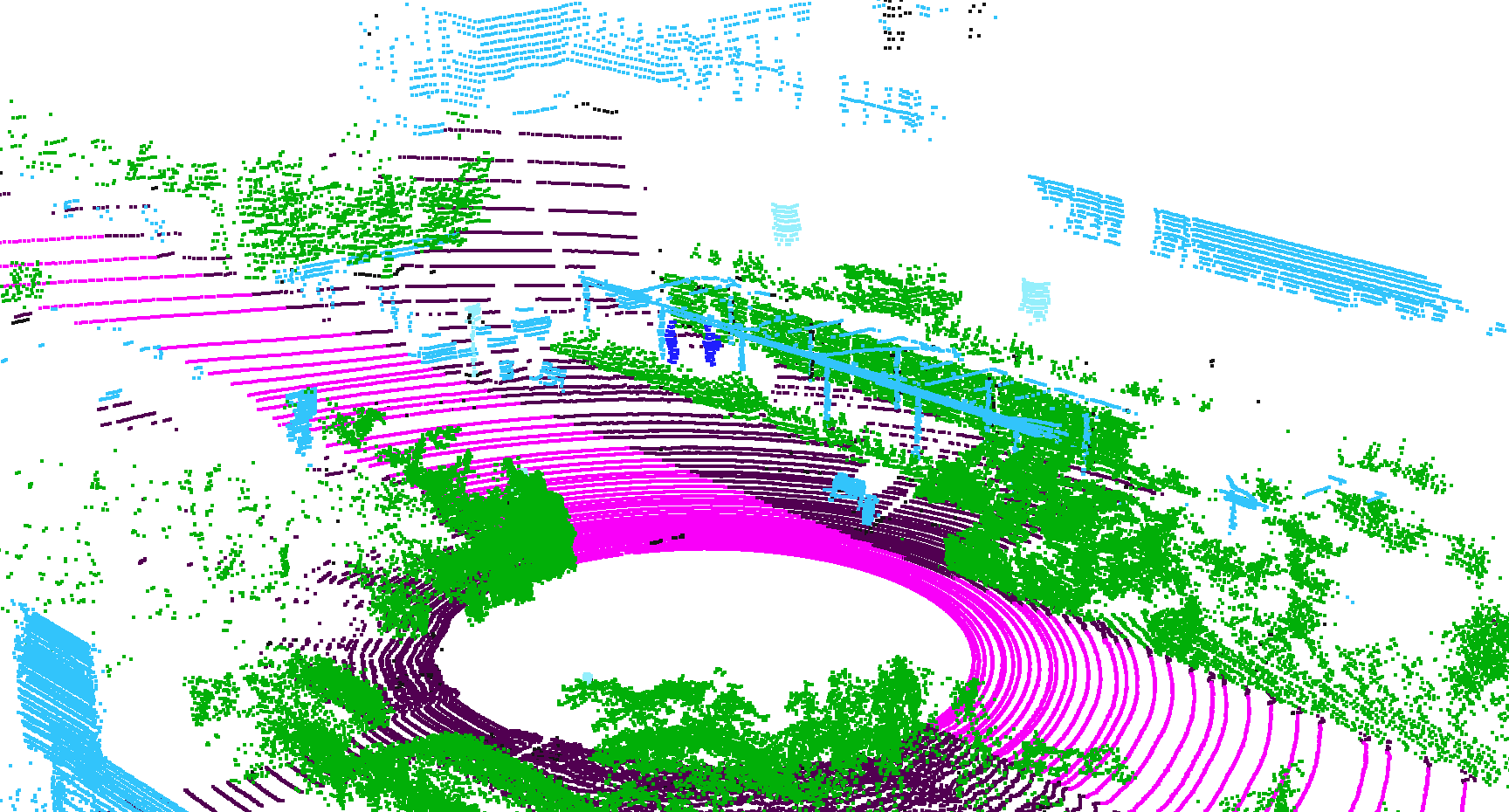}
    \caption{\textbf{First frame from sequence 15 of HelixNet.}~In this frame from the test set, we observe difficult situations with pedestrians under a urban infrastructure surrounded by vegetation.}
    \label{fig:bigzoom}
\end{figure}

\paragraph{Detailed results.} In the main paper, we showed that Helix4D yields state-of-the-art performance on HelixNet. \tabref{tab:confmat} shows the corresponding confusion matrix, which highlights three main sources of error in HelixNet. First, \emph{roads} and \emph{other-surfaces} are not necessarily easy to distinguish. Second, the model tends to misclassify \emph{traffic-signs}, \emph{pedestrians} and \emph{vegetation} as buildings. As urban hardscape such as post and poles, street fences, and public benchs being annotated as \emph{buildings}, \emph{traffic-signs} can be hard to identify. \emph{Pedestrians} can be inside of buildings or structures, making them hard to distinguish, see \figref{fig:bigzoom}. Third, the distinction between static and mobile cars can be difficult for vehicle stopped in the driving lanes. This requires both understanding the road configuration and the dynamic of the scene. See the attached HTML file for an interactive representation of some of these errors.

In \tabref{tab:detailed}, we report the quantitative values used in Figure~7 of the main paper, and include a comparison with the official performance of the trained models as reported by the papers they were introduced in.

\begin{table}[t]
    \centering
    \input{tikz/confmat_H4D-HNET}
    \caption{\textbf{Helix4D Semantic Segmentation of HelixNet.}~We report the confusion matrix of the predictions of Helix4D on the test set of HelixNet, evaluated in the online setting with slices of $72^\circ$.}
    \label{tab:confmat}
\end{table}

\begin{table*}[t]
    \centering
    \caption{\textbf{Semantic KITTI Validation Set Results.}~``Ours" stands for results as computed on our workstation using a NVIDIA TESLA V100 32Go GPU.}
    \label{tab:detailed}
    \renewcommand{\arraystretch}{1.25}
    \begin{tabular*}{\textwidth}{@{}l@{\extracolsep{\fill}}*{8}{c}@{}}
        \toprule
         & \multirow{1}{*}{Source} & Implementation & {\% scan} & \multicolumn{4}{c}{Inference time (ms.)} & \multirow{2}{*}{mIoU}\\
         & for performance& \& model weights & {time (ms.)} & min & mean & max & std & \\
        \midrule
        \multicolumn{4}{l}{\textbf{Cylinder3D~\cite{zhu2021cylindrical}}} & & \multicolumn{4}{r}{\scriptsize $55.9\times 10^6$ parameters} \\
        & \cite{zhu2021cylindrical} & --- & 100 (104) & ---  & --- & --- & ---   & 65.9 \\
        & Ours    & Official, pretrained & 100 (104) & 81 & 108 & 133 & 12 & 66.9  \\
        & Ours    & Official, pretrained & 50 (52)   & 56 & 74  & 94  & 7  & 66.2  \\
        & Ours    & Official, pretrained & 33 (35)   & 47 & 63  & 78  & 6  & 65.8  \\
        & Ours    & Official, pretrained & 25 (26)   & 46 & 57  & 71  & 5  & 65.7  \\
        & Ours    & Official, pretrained & 20 (21)   & 45 & 54  & 67  & 4  & 65.3  \\
        \midrule
        \multicolumn{4}{l}{\textbf{PolarNet~\cite{Zhang_2020_CVPR}}} & & \multicolumn{4}{r}{\scriptsize $13.6\times 10^6$ parameters} \\
        & \cite{Zhang_2020_CVPR} & --- & 100 (104) & --- & 94 & --- & --- & 53.6 \\ 
        & Ours    & Official, pretrained & 100 (104) & 37 & 49 & 51 & 4 & 58.2 \\ 
        & Ours    & Official, pretrained & 50 (52)   & 32 & 40 & 47 & 4 & 57.8  \\
        & Ours    & Official, pretrained & 33 (35)   & 31 & 41 & 47 & 3 & 57.1 \\ 
        & Ours    & Official, pretrained & 25 (26)   & 29 & 38 & 45 & 4 & 57.5 \\ 
        & Ours    & Official, pretrained & 20 (21)   & 29 & 36 & 49 & 5 & 56.9 \\ 
        \midrule
        \multicolumn{4}{l}{\textbf{SPVNAS~\cite{tang2020searching}}} & & \multicolumn{4}{r}{\scriptsize $10.8\times 10^6$ parameters} \\
              & \cite{tang2020searching} & --- & 100 (104) & --- & --- & --- & --- & 64.7  \\
         & Ours    & Official, pretrained & 100 (104) & 52 & 73 & 140 & 11 & 64.7  \\
         & Ours    & Official, pretrained & 50 (52)   & 44 & 52 & 100 & 4 & 62.5  \\
         & Ours    & Official, pretrained & 33 (35)   & 43 & 47 & 86  & 3 & 60.8  \\
         & Ours    & Official, pretrained & 25 (26)   & 41 & 45 & 70 & 3 & 59.0  \\
         & Ours    & Official, pretrained & 20 (21)   & 40 & 44 & 68  &  & 57.8  \\
        \midrule
        \multicolumn{4}{l}{\textbf{SalsaNeXt~\cite{cortinhal2020salsanext}}} & & \multicolumn{4}{r}{\scriptsize $6.7\times 10^6$ parameters} \\
         & Ours    & Official, pretrained & 100 (104) & 22 & 23 & 23 & 0.1 & 55.8  \\
         & Ours    & Official, pretrained & 50 (52)   & 12 & 13 & 16 & 0.2 & 55.8  \\
         & Ours    & Official, pretrained & 33 (35)   & 10 & 12 & 15 & 0.2 & 55.7  \\
         & Ours    & Official, pretrained & 25 (26)   & 9  & 10 & 13 & 0.2 & 55.6  \\
         & Ours    & Official, pretrained & 20 (21)   & 9  & 9.5 & 12 & 0.2 & 55.3 \\
        \bottomrule
    \end{tabular*}
\end{table*}

%% file: tikz/buckets.tex
\newcommand{\buckettitle}[1]{
    $\mathsf{bin}_#1\left(#1_u-#1_v\right)$
}
\newcommand{\bucketdelta}[2]{
    $#1_u-#1_v$ (#2.)
}
\newcommand{\bucketylabel}[1]{
    $\mathsf{bin}_#1\left(#1_u-#1_v\right)$
}

\begin{tikzpicture}
	\begin{axis}[
		title=\buckettitle{X},
		xlabel=\bucketdelta{X}{m},
		axis y line*=left,
		axis x line*=bottom,
		ytick ={-3,-2,-1,0, 1, 2, 3, 4, 5, 6, 7},
		xtick = {-6,-4,-2,0,2,4,6},
		xmin = -6, xmax=6,
		ymin = -3, ymax=3.1,
		legend pos=north west,
		width=.31\textwidth,height=.3\textwidth,
		ticklabel style={font=\tiny},
		every x tick/.style={black},
		every x tick label/.style={black},
		at={(0,0)}
	]
		
	\addplot[color=red, mark=none, style={thick}] coordinates {
		(-6, -3) (-3.79, -3)
		(-3.79, -2) (-2.23, -2)
		(-2.23, -1) (-0.75, -1)
		(-0.75, 0) (0.75, 0)
		(0.75, 1) (2.23, 1)
		(2.23, 2) (3.79, 2)
		(3.79, 3) (6, 3)
	};
	\end{axis}
\end{tikzpicture}
\hfill
\begin{tikzpicture}
	\begin{axis}[
		title=\buckettitle{Y},
		xlabel=\bucketdelta{Y}{m},
		axis y line*=left,
		axis x line*=bottom,
		ytick ={-3,-2,-1,0, 1, 2, 3, 4, 5, 6, 7},
		xtick = {-6,-4,-2,0,2,4,6},
		xmin = -6, xmax=6,
		ymin = -3, ymax=3.1,
		legend pos=north west,
		width=.31\textwidth,height=.3\textwidth,
		ticklabel style={font=\tiny},
		every x tick/.style={black},
		every x tick label/.style={black},
	]
		
	\addplot[color=red, mark=none, style={thick}] coordinates {
		(-6, -3) (-3.79, -3)
		(-3.79, -2) (-2.23, -2)
		(-2.23, -1) (-0.75, -1)
		(-0.75, 0) (0.75, 0)
		(0.75, 1) (2.23, 1)
		(2.23, 2) (3.79, 2)
		(3.79, 3) (6, 3)
	};
	\end{axis}
\end{tikzpicture}
\hfill
\begin{tikzpicture}
	\begin{axis}[
		title=\buckettitle{Z},
		xlabel=\bucketdelta{Z}{m},
		axis y line*=left,
		axis x line*=bottom,
		ytick ={-3,-2,-1,0, 1, 2, 3, 4, 5, 6, 7},
		xtick = {-6,-4,-2,0,2,4,6},
		xmin = -6, xmax=6,
		ymin = -3, ymax=3.1,
		legend pos=north west,
		width=.31\textwidth,height=.3\textwidth,
		ticklabel style={font=\tiny},
		every x tick/.style={black},
		every x tick label/.style={black},
	]
		
	\addplot[color=red, mark=none, style={thick}] coordinates {
		(-6, -3) (-1.25, -3)
		(-1.25, -2) (-0.75, -2)
		(-0.75, -1) (-0.25, -1)
		(-0.25, 0) (0.25, 0)
		(0.25, 1) (0.75, 1)
		(0.75, 2) (1.25, 2)
		(1.25, 3) (6, 3)
	};
	\end{axis}
\end{tikzpicture}
\hfill
\begin{tikzpicture}
	\begin{axis}[
		title=\buckettitle{T},
		xlabel=\bucketdelta{T}{s},
		axis y line*=left,
		axis x line*=bottom,
		ytick ={0, 1, 2, 3, 4, 5, 6, 7, 8},
		xtick = {-3,-2,-1, 0},
		xmin = -2.5, xmax=0,
		ymin = 0, ymax=2.1,
		legend pos=north west,
		width=.31\textwidth,height=.3\textwidth,
		ticklabel style={font=\tiny},
		every x tick/.style={black},
		every x tick label/.style={black},
	]
		\addplot[color=red, mark=none, style={thick}] coordinates {
		(-2.5, 0) (-1.5, 0)
		(-1.5, 1) (-0.5, 1)
		(-0.5, 2) (0, 2)
		};
	\end{axis}
\end{tikzpicture}

%% file: tikz/heads.tex
\newcommand{\headtitle}[1]{
    $\hat{y}_{h,#1}$
}
\newcommand{\headdelta}[2]{
    $\mathsf{bin}_#1$
}
\newcommand{\addheadplot}[3]{
    \addplot[color=#3, mark=none, style={thick}] table [x={b},y={h#2}] {results/heads#1.txt};\label{head:#2}
}
\newcommand{\addheadylabel}[1]{
    $\sum_\cV \left(\key_v^h\right)^\intercal \left(\posenc^h(u,v)\right)$
}

\newcommand{\addheaddimplot}[4]{
    \begin{tikzpicture}
    	\begin{axis}[
    		title=\headtitle{#1},
    		xlabel=\headdelta{#1}{bin},
    		axis y line*=left,
    		axis x line*=bottom,
    		xtick = #4,
    		xmin = #2, xmax=#3,
    		width=.29\textwidth,height=.3\textwidth,
    		ticklabel style={font=\tiny},
    		every x tick/.style={black},
    		every x tick label/.style={black},
    	]
    	
    	\addheadplot{#1}{1}{blue}
    	\addheadplot{#1}{2}{orange}
    	\addheadplot{#1}{3}{green}
    	\addheadplot{#1}{4}{red}
    	
    	\end{axis}
    \end{tikzpicture}
}
\addheaddimplot{X}{-3}{3}{{-3,-2,-1,0,1,2,3}}
\hfill
\addheaddimplot{Y}{-3}{3}{{-3,-2,-1,0,1,2,3}}
\hfill
\addheaddimplot{Z}{-3}{3}{{-3,-2,-1,0,1,2,3}}
\hfill
\addheaddimplot{T}{0}{2}{{0,1,2}}

\begin{tikzpicture}
	\node[draw,minimum width=.75\textwidth]{
	\begin{tabular}[width=\textwidth]{c}
        ~~~~~
        {\ref{head:1}~{Head 1}}~~~~~
        {\ref{head:2}~{Head 2}}~~~~~
        {\ref{head:3}~{Head 3}}~~~~~
        {\ref{head:4}~{Head 4}}~~~~~
    \end{tabular}
	};
\end{tikzpicture}
\vspace{.5em}

%% file: tikz/confmat_H4D-HNET.tex

\def\colorModel{hsb}

\newcommand\ColCellRORO[1]{
  \pgfmathparse{#1<50?1:0}  
    \ifnum\pgfmathresult=0\relax\color{white}\fi
  \pgfmathsetmacro\compA{.6}      
  \pgfmathsetmacro\compB{#1/100} 
  \pgfmathsetmacro\compC{1-.1*#1/100}      
  \edef\x{\noexpand\centering\noexpand\cellcolor[\colorModel]{\compA,\compB,\compC}}\x #1
  } 

\newcommand\nclasses{9}

\newcommand{\hnetconfmatclasses}[1]{\multicolumn{1}{p{.85cm}}{{\rotatebox{35}{#1}}}}
\newcommand{\PreserveBackslash}[1]{\let\temp=\\#1\let\\=\temp}
\newcolumntype{C}[1]{>{\PreserveBackslash\centering}p{0.85cm}}
\newcolumntype{R}[1]{>{\PreserveBackslash\raggedleft}p{#1}}
\newcolumntype{L}[1]{>{\PreserveBackslash\raggedright}p{#1}}
\newcolumntype{E}{>{\collectcell\ColCellRORO}m{.85cm}<{\endcollectcell}}

\newcommand\dohhline{}
    
\arrayrulecolor{white}
\renewcommand{\arraystretch}{1.25}
\begin{tabular}{@{}p{.5cm}r*{\nclasses}{|E}|@{}}
    \multicolumn{1}{c}{} & \multicolumn{1}{c}{} & \multicolumn{\nclasses}{c}{\textbf{Predicted label}} \\
    \multicolumn{1}{c}{} & \multicolumn{1}{c}{} & \hnetconfmatclasses{Road} & \hnetconfmatclasses{Other surface} & \hnetconfmatclasses{Building} & \hnetconfmatclasses{Vegetation} & \hnetconfmatclasses{Traffic signs} & \hnetconfmatclasses{Static vehicle} & \hnetconfmatclasses{Moving vehicle} & \hnetconfmatclasses{Pedestrian} & \hnetconfmatclasses{Artifact}\\ \dohhline
    \multirow{\nclasses}{*}{\rotatebox{90}{\textbf{True label}}}
    & Road           & 94.4 &5.6 & 0.0  & 0.0 & 0.0 & 0.0 & 0.0 & 0.0 & 0.0 \\ \dohhline
    & Other surface  & 6.8 & 90.0 & 1.4 & 1.8 & 0.0 & 0.0 & 0.0 & 0.0 & 0.0   \\ \dohhline
    & Building       & 0.0 & 0.7 & 98.2 & 0.7 & 0.2 & 0.1 & 0.0 & 0.1 & 0.0   \\ \dohhline
    & Vegetation     & 0.1 & 3.4 & 6.8 & 88.9 & 0.6 & 0.2 & 0.0 & 0.1 & 0.0   \\ \dohhline
    & Traffic signs  & 0.0 & 0.5 & 17.9 & 4.7 & 76.1 & 0.4 & 0.1 & 0.3 & 0.0   \\ \dohhline
    & Static vehicle & 0.3 & 0.4 & 1.2 & 0.3 & 0.1 & 75.8 & 21.7 & 0.3 & 0.0   \\ \dohhline
    & Moving vehicle & 0.3 & 0.2 & 2.7 & 0.2 & 0.2 & 12.6 & 83.5 & 0.4 & 0.0   \\ \dohhline
    & Pedestrian     & 0.2 & 1.3 & 8.1 & 1.0 & 0.4 & 1.3 & 0.7 & 87.1 & 0.0 \\ \dohhline
    & Artifact       & 0.5 & 2.7 & 0.6 & 1.5 & 0.2 & 0.4 & 0.1 & 0.0 & 94.0   \\ \dohhline
\end{tabular}
\arrayrulecolor{black}

\vspace{.5em}